\documentclass[10pt,twocolumn,letterpaper]{article}

\usepackage{iccv}              

\newcommand{\red}[1]{{\color{red}#1}}

\newcommand{\ao}[1]{{\color{blue}#1}}

\def\figvspacetop{\vspace{-10pt}}
\def\figvspace{\vspace{-8pt}}

\usepackage[pagebackref,breaklinks,colorlinks]{hyperref}

\usepackage{makecell}

\def\paperID{10822} 
\def\confName{ICCV}
\def\confYear{2025}
\usepackage{tcolorbox}
\usepackage{listings}
\usepackage{amsmath,amssymb}
\usepackage{algorithmic}
\usepackage{algorithm}
\usepackage{bm}
\usepackage{booktabs}
\usepackage{comment}
\usepackage{multirow}
\usepackage{framed}
\usepackage{tabularx}

\def\red#1{{\textcolor{red}{#1}}}

\def\magenta#1{{\textcolor{magenta}{#1}}}

\usepackage{url}
\makeatletter
\newcommand{\figcaption}[1]{\def\@captype{figure}\caption{#1}}
\newcommand{\tblcaption}[1]{\def\@captype{table}\caption{#1}}
\makeatother

\newcount\K
\def\bla#1{
\K=0 \loop\ifnum\K<#1
{\textcolor[gray]{0.9}{{\it bla bla bla bla bla bla bla bla bla bla bla bla bla bla bla}}}
\advance\K by1\repeat
}

\title{GeoProg3D: Compositional Visual Reasoning for \\ 
City-Scale 3D Language Fields}

\author{
Shunsuke Yasuki\textsuperscript{1} \quad
Taiki Miyanishi\textsuperscript{2,3} \quad
Nakamasa Inoue\textsuperscript{4} \quad
Shuhei Kurita\textsuperscript{5,6} \\
Koya Sakamoto\textsuperscript{2} \quad
Daichi Azuma\textsuperscript{2,7} \quad
Masato Taki\textsuperscript{1} \quad
Yutaka Matsuo\textsuperscript{2}
\\[2ex]
\textsuperscript{1}Rikkyo University \quad
\textsuperscript{2}University of Tokyo \quad
\textsuperscript{3}ATR \quad
\textsuperscript{4}Institute of Science Tokyo \\
\textsuperscript{5}National Institute of Informatics \quad
\textsuperscript{6}NII LLMC \quad
\textsuperscript{7}Sony Semiconductor Solutions \quad
\\[2ex]\\
{\tt\small Project Page: \href{https://snskysk.github.io/GeoProg3D/}{https://snskysk.github.io/GeoProg3D/}}
}

\def\methodname{GeoProg3D\xspace}
\def\datasetname{GeoEval3D\xspace}

\begin{document}
\makeatletter
\g@addto@macro\@maketitle{
\begin{figure}[H]
\setlength{\linewidth}{\textwidth}
\setlength{\hsize}{\textwidth}
\centering
\vspace{-10pt}
\includegraphics[width=\linewidth]{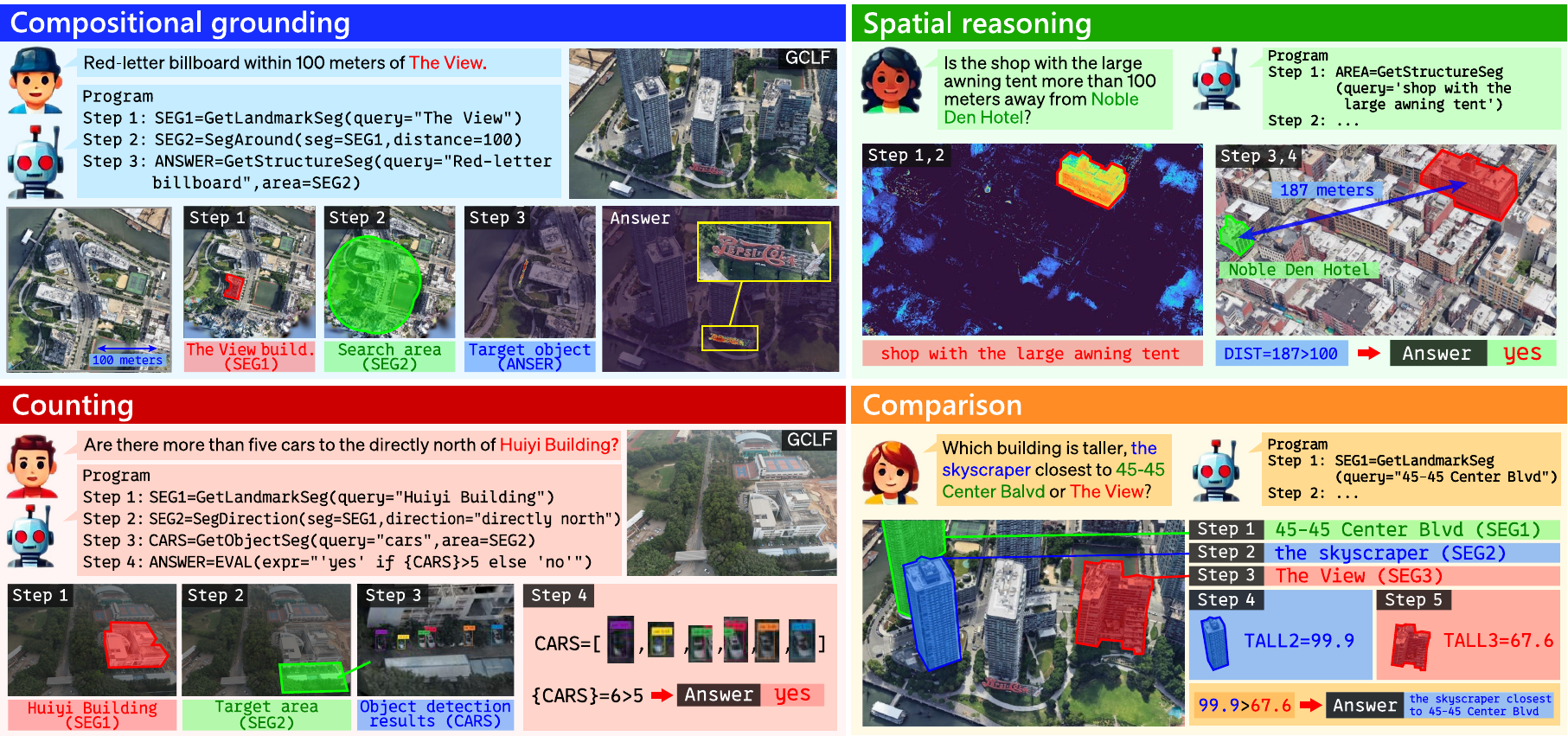}
%

\vspace{-0.5em}
\caption{
Overview of the proposed compositional geographic reasoning task. 
This task enables natural language interaction with city-scale 3D scenes, supporting diverse geographic reasoning scenarios.
Given a natural language query, our {\methodname} decomposes the task into modular steps, executes structured visual programs using geographic APIs, and integrates results 
for interpretable and accurate reasoning.
}

\label{fig:teaser}

\end{figure}
}
\makeatother

\maketitle

\begin{abstract}
The advancement of 3D language fields has enabled intuitive interactions with 3D scenes via natural language. 
However, existing approaches are typically limited to small-scale environments, lacking the scalability and compositional reasoning capabilities necessary for large, complex urban settings. 
To overcome these limitations, we propose \methodname, a visual programming framework that enables natural language-driven interactions with city-scale high-fidelity 3D scenes. 
\methodname consists of two key components: (i) a Geography-aware City-scale 3D Language Field (GCLF) that leverages a memory-efficient hierarchical 3D model to handle large-scale data, integrated with geographic information for efficiently filtering vast urban spaces using directional cues, 
distance measurements, elevation data, and landmark references;
and (ii) Geographical Vision APIs (GV-APIs), specialized geographic vision tools such as area segmentation and object detection.
Our framework employs large language models (LLMs) as reasoning engines to dynamically combine GV-APIs and operate GCLF, effectively supporting diverse geographic vision tasks. 
To assess performance in city-scale reasoning, we introduce \datasetname, a comprehensive benchmark dataset containing 952 query-answer pairs across five challenging tasks:
grounding, spatial reasoning, comparison, counting, and measurement. 
Experiments demonstrate that \methodname significantly outperforms existing 3D language fields and vision-language models across multiple tasks.
To our knowledge, \methodname is the first framework enabling compositional geographic reasoning in high-fidelity city-scale 3D environments via natural language.

\end{abstract}
\vspace{-10pt}

\section{Introduction}

Large-scale 3D scene reconstruction has emerged as a pivotal technology enabling a wide spectrum of real-world applications. 
These include the creation of 3D digital worlds~\cite{gu2023uenerf,li2023matrixcity,xie2024citydreamer}, urban scene editing~\cite{chen2024stylecity,xie2024gaussiancity,xie2024citydreamer}, and autonomous driving simulation~\cite{guo2023streetsurf,Yang_2023_CVPR,yuan2024presight}.
In particular, recent advances in radiance fields such as Neural Radiance Fields (NeRF) ~\cite{mildenhall2020nerf} and 3D Gaussian Splatting (3D-GS)~\cite{kerbl3Dgaussians} have revolutionized the ability to reconstruct city-scale 3D models with unprecedented fidelity.
These methods leverage diverse data sources, from street-level imagery from vehicles~\cite{blocknerf,zhou2024drivinggaussian,fischer2024dynamic} to aerial and satellite photography~\cite{liu2024citygaussian,shuaiLoG2024,xu2023gridguided,song2024city,liu2025citygaussianv2}, enabling the high-fidelity representations of entire cities.
However, intuitive and efficient interaction with these detailed 3D city models using natural language remains largely unexplored.


Recently, a promising development in this direction has been the emergence of 3D language fields, which facilitate natural language interaction with high-resolution 3D scenes, enabling precise localization of specific objects or regions within these scenes~\cite{kerr2023LERF,qin2023LangSplat,shi2024language,Bhalgat2024N2F2,ji2025fastlgs}.
However, when extending conventional 3D language fields for large-scale urban 3D scenes, two fundamental difficulties emerge:
(1) \textit{Scalability for city-scale 3D data}:
Since existing methods primarily focus on indoor scenes, the high-fidelity reconstruction and efficient handling of large-scale urban scenes exceeding 1$km^2$~\cite{Lin2022UrbanScene3D} remain challenging.
(2) \textit{Enhanced task versatility for urban applications}:
Current 3D language fields predominantly focus on localizing discrete objects through language queries,
yet they fail to adequately address the multifaceted demands intrinsic to urban applications such as interpreting spatial relationships, quantifying object numbers and sizes, or recognizing landmark identifiers.
Therefore, extending 3D language fields to urban-scale environments necessitates a more scalable and compositional framework capable of comprehending the intricate complexities of vast citywide landscapes.



In this work, we address these difficulties by proposing GeoProg3D, a visual programming framework which consists of two key components: (i) a Geography-aware City-scale 3D Language Field (GCLF), and (ii) Geographical Vision APIs (GV-APIs) combined with LLMs for code generation.
To address the scalability issue (1), GCLF utilizes memory-efficient hierarchical 3D Gaussians to represent 3D language fields. The structure is useful for high-fidelity reconstruction and fast inference of vast urban environments.
It also enables localization of objects based on landmark names by aligning the linguistic and geographic information of real-world 2D maps with 3D Gaussians.
In response to issue (2) and to enhance versatility, GV-APIs provide a set of image and geographical processing APIs, including object detection, area segmentation, and distance measurement on GCLF. 
Figure~\ref{fig:teaser} provides an overview of the response process of \methodname across four distinct scenarios.
By leveraging large language models (LLMs) as reasoning engines to dynamically integrate GV-APIs, GeoProg3D effectively decomposes complex queries into simpler subtasks that can be processed on GCLF, facilitating compositional reasoning over city-scale 3D data.

To validate the effectiveness, we introduce novel tasks designed to assess urban-scale geographic visual reasoning capabilities and present \datasetname, a benchmark dataset specifically developed for this task.
\datasetname encompasses 952 carefully designed query-answer pairs across five essential within realistic urban environments derived from city-scale 3D reconstructions~\cite{xie2024citydreamer,Lin2022UrbanScene3D}.
The dataset covers an area exceeding 3 $km^2$ across diverse urban settings in New York (U.S.) and Shenzhen (China). 

In summary, our contributions are threefold:
\begin{itemize} 
\item We propose \methodname, a framework for compositional reasoning over city-scale 3D language fields, where visual programming can perform various 3D geographic vision tasks via image and geographic APIs.
\item We introduce five geographic compositional reasoning tasks in city-scale 3D scenes through natural language queries, and present a new benchmark dataset, \datasetname, that encompasses these challenges.
\item We demonstrate the superior performance of \methodname on \datasetname in comparison to existing 3D language fields and state-of-the-art VLMs by a significant margin. 
\end{itemize}


\if[]

\item \magenta{We introduce the task of compositional geographic reasoning over city-scale, high-resolution 3D scenes using natural language queries.}
\item \magenta{We propose \methodname, a framework for compositional reasoning over city-scale 3D language fields, where visual programming can perform various 3D geographic vision tasks via image and geographic APIs.}
\item We release \datasetname, a new benchmark dataset for city-scale 3D scene understanding, which covers five geographical vision tasks. 
\item We demonstrate the superior performance of \methodname on \datasetname in comparison to existing 3D language fields \ao{and state-of-the-art VLMs} by a significant margin. 
\begin{itemize} 
\item \magenta{We introduce the task of compositional geographic reasoning over city-scale, high-resolution 3D scenes using natural language queries.}
\item \magenta{We present \methodname, a framework for compositional reasoning over city-scale 3D language fields, where visual programming can perform various 3D geographic vision tasks via image and geographic APIs.}
\item We release \datasetname, a new benchmark dataset for city-scale 3D scene understanding, which covers five geographical vision tasks. 
\item We demonstrate the superior performance of \methodname on \datasetname in comparison to existing 3D language fields \ao{and state-of-the-art VLMs} by a significant margin. 
\end{itemize}
\fi


\section{Related Work}
\noindent
\textbf{City-scale 3D scene reconstruction.}
3D reconstruction from large-scale image collections has garnered considerable attention and achieved significant advancements in recent years. 
NeRF have been widely employed to produce high-fidelity 3D representations from 2D image inputs by modeling volumetric scenes~\cite{blocknerf,Turki_CVPR2022_MegaNERF,xiangli2022bungeenerf,zhang2024aerial,xu2023gridguided,Zhang2023Efficient,li2024nerfxl}. 
However, the unique nature of volume rendering in NeRF leads to substantial resource requirements,
particularly for high-resolution outputs.
More recently, city-scale 3D scene reconstruction using 3D-GS has emerged as a highly efficient alternative to NeRF-based approaches, offering improvements in rendering speed, scalability, and adaptability to real-time applications~\cite{liu2024citygaussian,shuaiLoG2024,liu2024citygaussianv2,Lin_2024_CVPR,zhang2024garfield,song2024city,ham2024dragon,yuchen2024dogaussian,KerblTOG2024higs}. 
Despite these advancements in precise 3D reconstruction, the development of technologies to ground geolocation text within urban scenes and retrieve geolocation information remains underexplored.
We propose GeoProg3D, a novel method designed to ground textual queries and instructions onto precise city-scale scenes for comprehensive analyses.

\vspace{3pt} 
\noindent \textbf{3D language field.}
Following the rapid advancement of 3D representations, grounding 3D data to text has become an intensive research topic in recent years.
Classical approaches to text grounding in point clouds include ScanRefer~\cite{chen2020scanrefer}, which uses textual descriptions for 3D localization to generate 3D bounding boxes, and ScanQA~\cite{azuma2022scanqa}, which addresses 3D question answering using point clouds from the ScanNet dataset~\cite{dai2017scannet, yeshwanthliu2023scannetpp}.
Neural sematic fields~\cite{pmlr-v164-blukis22a,Shafiullah2022CLIPFieldsWS}, NeRF decomposition~\cite{DeRF,decompnerf,kobayashi2022distilledfeaturefields} 
 and 3D LLMs~\cite{3dllm,tang2024minigpt,xu2024pointllm,3d-vista} have been proposed to enhance spatial information representation.
LERF~\cite{kerr2023LERF} embeds CLIP features into a 3D scene representation constructed with NeRF, enabling 3D spatial search using open vocabulary queries. 
However, as NeRF-based approaches have limitations in both speed and accuracy, 3D-GS based language embed models of LangSplat~\cite{qin2023LangSplat} and LEGaussians~\cite{shi2024language} and their applications emerge~\cite{Chen2024PVLFF,ji2024-graspsplats,zhang20243DitScene}.
Specifically, LangSplat avoids the rendering cost problem by adopting 3D-GS for scene representation, and constructs 3D language fields with compressed CLIP features into 3D Gaussians.
CLIP features are embedded as features with 3D clear boundaries, guided by the 2D object masks obtained by applying SAM~\cite{kirillov2023segment} to multiple training views.
These 3D language fields are primarily designed for finite-scale indoor scenes. To the best of our knowledge, there are no city-scale 3D language fields.

\vspace{3pt} 
\noindent \textbf{LLMs for visual reasoning tasks.}
Recent advancements in LLMs and Vision-Language Models (VLMs) have enabled the composition of program codes for extracting relevant features from texts and images, summarizing them, and executing symbolic reasoning tasks~\cite{Gupta2022VisProg,Suris2023ViperGPT,lu2023chameleon,feng2023layoutgpt,Subramanian2023CodeVQA,Schick2023ToolformerLM}. 
Visual Programming~\cite{Gupta2022VisProg} facilitates the resolution of compositional vision tasks by receiving natural language queries generating executable Python programs that invoke external functions. 
ViperGPT~\cite{Suris2023ViperGPT} enables Python code generation and execution for visual reasoning and question answering. 
CodeVQA~\cite{Subramanian2023CodeVQA} also introduces a code generation approach tailored for positional question answering from images.
In the domain of 3D visual grounding, systems have been developed for zero-shot open-vocabulary 3D visual grounding, primarily within narrow indoor scenes~\cite{Subramanian2023CodeVQA}, intending for narrow indoor scenes, and the versatility of visual programming is only utilized for the grounding of 3D point cloud space.
In contrast, our work introduces GeoProg3D, a visual programming system designed to handle complex queries within city-scale 3D spaces.

\vspace{3pt} 
\noindent \textbf{Geography-aware vision and language.}
VLMs also have significantly propelled research in geography-aware vision and language modelings, enabling to understand and reason about visual data within a geographical context~\cite{zhou2024vlgfm}.
A diverse array of geographical tasks has been proposed, encompassing geo-localization~\cite{vivanco2023geoclip,klemmer2023satclip,liu2024remoteclip,Haas_CVPR2024_PIGEON,xu2024addressclip}, visual grounding~\cite{Yuan2022TGRS,shu2022rsvg}, image captioning~\cite{qu2016cits,Lu2018ExploringMA,Meng2024MultiscaleGT,zhang2024earthmarker}, question answering~\cite{Zheng2021RSIVQA,Chappuis2022CVPR}, and text-to-image generation~\cite{khanna2024diffusionsat,tang2024crsdiff}. 
Recently, Vision-Language Grounding Foundation Models (VLGFMs) have been proposed as a framework for solving these multiple challenges simultaneously by fine-tuning VLMs~\cite{zhang2024earthgpt,Hu2023RSGPTAR,Kuckreja_2024_CVPR,wang2023skyscript,dilxat2024lhrs,luo2024sky,Zhang2024RS5M,irvin2024teochat,pang2024vhmversatilehonestvision}, enhancing their capacity to understand and analyze complex geospatial information.
However, VLGFMs are limited to handling only 2D top-down view images and are not capable of processing city-scale 3D data.

\begin{figure*}
\vspace{-1.5em}
\centering
\includegraphics[width=\linewidth]{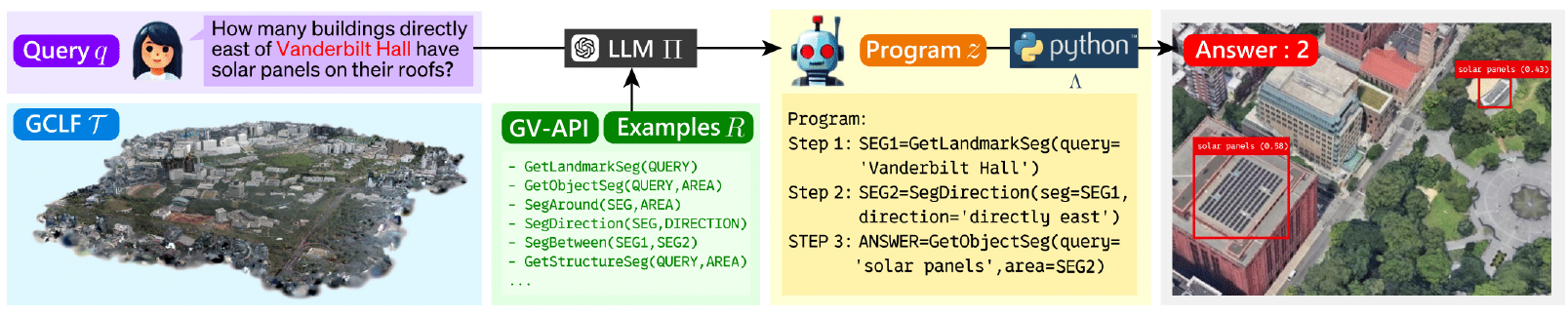}
\vspace{-2.4em}
\caption{\textbf{Framework overview.}
Given a user query, GeoProg3D generates a visual program via LLM in-context learning. The program operates GCLF by combining Geographical Vision APIs (GV-APIs)  and answers the query.
}
\label{fig:geoprog3d_framework}
\vspace{-1em}
\end{figure*}

\begin{figure*}
\vspace{-0.1em}
\centering
\includegraphics[width=\linewidth]{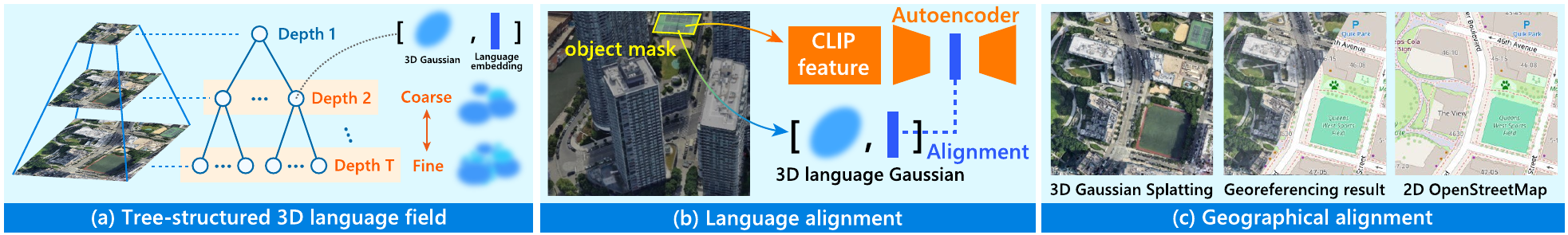}
\vspace{-2em}
\caption{\textbf{GCLF structure.} (a) Coarse-to-fine tree structure to represent 3D scenes. Each node represents a pair of a 3D Gaussian and a language embedding. (b) Language alignment using CLIP features. (c) Geographical alignment using OpenStreetMap.}
\label{fig:gclf}
\vspace{-1.2em}
\end{figure*}

More closely related to our work, 
recent studies have introduced 3D vision-language tasks that focus on 3D point cloud data of urban environments, moving beyond the reliance on top-down 2D city images.
Notable examples include include 3D visual grounding~\cite{Kolmet_2022_CVPR,xia2024text2loc,miyanishi2023cityrefer} and 3D question answering~\cite{sun20243d}. 
Early work explores the use of VLMs for predicting urban attributes from 3D city models~\cite{opencity3d2025}.
However, these frameworks are typically limited to single 3D vision tasks or have not been extended to high-resolution, city-scale 3D reconstruction models.
In contrast, our approach is designed for high-fidelity, city-scale 3D data and supports a diverse array of tasks, including compositional grounding, counting, and spatial reasoning, all of which require compositional reasoning.
This versatility enables comprehensive interaction with complex urban environments through natural language.

\section{GeoProg3D Framework}
\label{sec:method}


\if[]
This section introduces GeoProg3D, a framework that enables users to interact with large-scale 3D scenes using natural language queries.
As illustrated in Figure~\ref{fig:teaser}, our framework supports multiple tasks, including grounding and spatial reasoning, through program generation that facilitates compositional 3D scene understanding.
Our approach differs from previous visual programming methods~\cite{Suris2023ViperGPT,Gupta2022VisProg} in two key aspects.
First, GeoProg3D can handle queries on 3D scenes instead of 2D images. 
To achieve this, we build a GCLF, a city-scale 3D language field geographically aligned with real-world 2D maps. 
Second, GeoProg3D can process queries related to geographic information through Geographical Vision APIs (GV-APIs) that manipulate the GCLF to produce responses in a compositional manner.
We presents a framework overview in Figure~\ref{fig:geoprog3d_framework}, which processes the input query $q$ in two steps: program generation and execution.
In the first step, a program $z \in \mathcal{Z}$ that answers the query is generated as $z = \Pi(q, R)$, where $\Pi$ is an LLM, $R$ is in-context examples using GV-APIs, and $\mathcal{Z}$ is the set of executable Python programs.
In the second step, the generated program $z$ is executed to obtain the answer $a$ as $a = \Lambda(z; \mathcal{T})$, where $\Lambda$ is the Python execution engine and $\mathcal{T}$ is a GCLF. 
\ao{Below, we describe the details of the GCLF, GV-APIs, visual programming in ~\cref{sec:method:3d-fields,sec:method:geo-vision-apis,sec:method:visprog}.}
\fi

This section introduces GeoProg3D, a framework that enables users to interact with large-scale 3D scenes using natural language queries
, consisting of two important components: (i) Geography-aware City-scale 3D Language Field (GCLF), and (ii) Geographical Vision APIs (GV-APIs), specialized modules for geographic visual reasoning tasks.
GCLF extends 3D language fields to city-scale and localizes objects, regions, and landmarks using 
natural language queries.
GeoProg3D utilizes visual programming via LLMs to dynamically combine GV-APIs for operating GCLF, and is adaptable to 
a variety of 3D geographic vision tasks.

\subsection{Overview of GeoProg3D}
We present an overview of GeoProg3D framework in Figure~\ref{fig:geoprog3d_framework}, which consists of two steps: the generation and execution of visual programs.
In the first step, a program $z \in \mathcal{Z}$ that answers the query $q$ is generated as $z = \Pi(q, R)$, where $\Pi$ is an LLM, $R$ is in-context examples using GV-APIs, and $\mathcal{Z}$ is the set of executable Python programs (GV-APIs).
In the second step, the generated program $z$ is executed to obtain the answer $a$ as $a = \Lambda(z; \mathcal{T})$, where $\Lambda$ is the Python execution engine and $\mathcal{T}$ is GCLF. 
The key differences between GeoProg3D and existing visual programming methods~\cite{Suris2023ViperGPT,Gupta2022VisProg} are that GeoProg3D can handle 3D scene queries rather than 2D images, and this enables operations and control over 3D language fields.
Below, we describe the details of GCLF, GV-APIs, visual programming in ~\cref{sec:method:3d-fields,sec:method:geo-vision-apis,sec:method:visprog}.

\subsection{\scalebox{0.955}[1]{Geography-aware City-scale 3D Language Fields}}
\label{sec:method:3d-fields}


\noindent \textbf{Scene representation.}
As a first step toward constructing GeoProg3D, we design a city-scale 3D language field (GCLF) with two key features: high fidelity and fast inference. 
City-scale reconstructions often produce rough details, which can negatively impact localization and image processing performance. 
Additionally, fast rendering is essential for practical use of large-scale 3D city models. 
To meet these requirements, GCLF represents urban scenes by embedding language into tree-structured 3D Gaussians. 
As described in~\cite{shuaiLoG2024}, this tree structure learns the nested relationships between Gaussians for detailed representation and those for an overview across multiple layers.
%
%
Figure~\ref{fig:gclf}\red{a} shows this structure. 
The rendering algorithm dynamically selects a hierarchical level in which the 3D Gaussian has a diameter of less than one pixel in image space, and efficiently renders the space that is far from the viewpoint.

\begin{figure}[ht]
    \centering
    \includegraphics[width=1\linewidth]{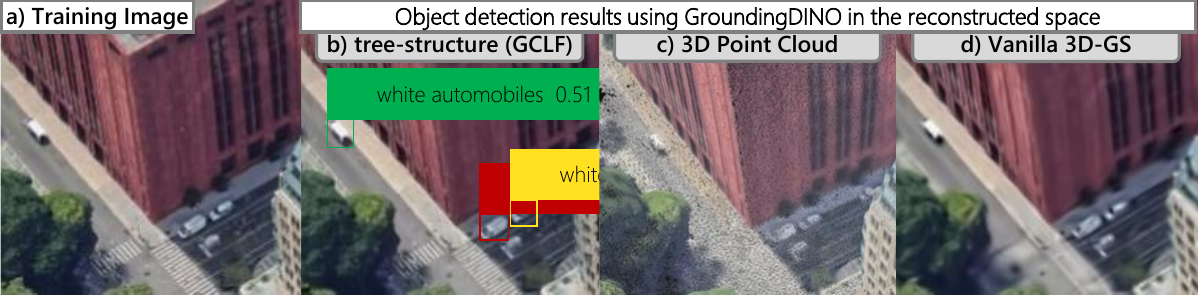}
    \vspace{-2em}
    \caption{ 
    The results of object detection using GroundingDINO.
    }
    \label{fig:fig_comp_pointcloud_vanilla}
    \vspace{-0.6em}
\end{figure}

\noindent \textbf{Why GCLF?}
To demonstrate the advantages of GCLF over conventional 3D Gaussian Splatting (3D-GS), we compare its efficiency and reconstruction quality. 
As shown in Table~\ref{tab:comparison_3dgs_log_stats}, GCLF requires tens of times more 3D Gaussians than LangSplat (vanilla 3D-GS) due to its hierarchical reconstruction approach, yet the rendering speed increases only by a few times, ensuring high-speed rendering remains feasible~\cite{shuaiLoG2024}. 
Furthermore, as illustrated in Figure~\ref{fig:fig_comp_pointcloud_vanilla}, the GCLF tree structure reconstructs city scenes with higher fidelity than both point clouds and vanilla 3D-GS, enabling successful inferences such as object detection on rendered images.

\noindent \textbf{Geo-visual integration.}
To effectively integratevisual and geographic data, 
we train our language embeddings to align with CLIP image features~\cite{radford2021learning} (see Figure~\ref{fig:gclf}\red{b}), thereby extending the LangSplat approach~\cite{qin2023LangSplat} into a tree-structured 3D language field. 
Additionally, we georeference the Gaussian coordinates to real-world coordinates (see Figure~\ref{fig:gclf}\red{c}), which enables our system to generate precise responses incorporating geographic details such as landmark names and measurements in real-world units.

\begin{table}[t]
\centering
\setlength{\tabcolsep}{5pt}
\scriptsize
\begin{tabular}{lccccc}
\toprule
& \multicolumn{2}{c}{\# Gaussians} & \multicolumn{2}{c}{Rendering speed (ms)}  \\
\cmidrule{2-3} \cmidrule{4-5}
{Scene} & LangSplat & GCLF & LangSplat & GCLF \\
\midrule
Center Blvd     & 37,212     & 1,136,015    & 2.73  & 14.46  \\
World Fin Ctr   & 30,278     & 763,432      & 2.21  & 7.99   \\
Mott St         & 33,846     & 1,253,668    & 2.97  & 14.48  \\
Washington Sq   & 31,757     & 950,932      & 2.34  & 12.35  \\
UrbanScene3D    & OOM        & 37,813,418   & OOM   & 20.83  \\
\bottomrule
\end{tabular}
\vspace{-1em}
\caption{Number of Gaussians and inference speed.}
\label{tab:comparison_3dgs_log_stats}
\vspace{-1em}
\end{table}

\begin{figure*}
\vspace{-1.5em}
\centering
\includegraphics[width=\linewidth]{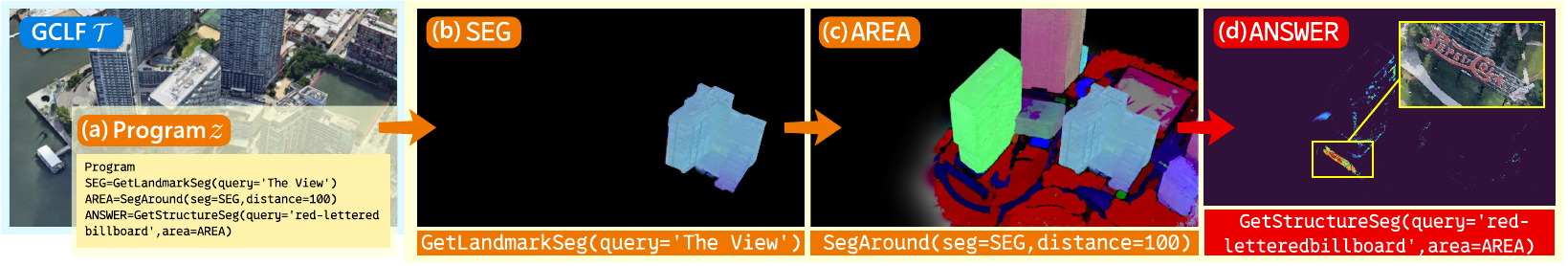}
\figvspacetop
\vspace{-1.1em}
\caption{\textbf{Execution example.} (a) Program code generated from the query ``Red-letter billboard within 100 meters of \textit{The View}.'' that consists of three steps. (b) \texttt{GetLandmarkSeg} identifies the building \textit{The View}. (c) \texttt{SegAround} retrieves the area around \textit{The View} within a 100-meter radius. (d) \texttt{GetStructureSeg} produces the segment where the confidence map from GCLF is visualized.}
\label{fig:exec}
\vspace{-1em}
\end{figure*}

\noindent \textbf{Training.}
%
Given a set of multi-view images $\mathcal{D} = \{x_{i}\}_{i=1}^{N}$ for training, a GCLF is trained in the following three steps.
First, coarse vanilla 3D Gaussian primitives $G = \{\bm{g}_{j}\}_{j=1}^{K}$ are trained on $\mathcal{D}$
, where $\bm{g}_{j}$ is a vector representing the position, colors, scale and rotation of the $j$-th Gaussian and $K$ is the number of Gaussians.
Next, following the learning method in \cite{shuaiLoG2024}, the tree-structure $G^{\prime}$ is trained based on $G$.
Third, language embedding into 3D Gaussian is trained.
Unlike LangSplat~\cite{qin2023LangSplat}, which is based on vanilla 3D-GS, the embedding is trained on the tree structure $G^{\prime}$ using our unique implementation.
As shown in Figure~\ref{fig:gclf}\red{b}, the language embedding teacher data $\bm{c}^{(m)}$ is obtained by applying a trained autoencoder for each scene to CLIP features, in the same way as LangSplat\cite{qin2023LangSplat}.
Here, $c$ is the CLIP image encoder and $m$ is the index of the object mask obtained from the training image using the Segment Anything model~\cite{Kirillov_2023_ICCV}.

\noindent \textbf{Georeferencing.}
After training GCLF $\mathcal{T}$, georeferencing is performed to align the 3D scene with a 2D map in a semi-automatic manner.
First, top-down view images of four small areas are rendered, each with an image size of 1024$\times$1024 pixels.
Second, more than 20 landmark points are manually chosen as points of interest that are visible in OpenStreetMap.
Finally, the geometric transformation is computed between the real-world coordinates on the map and the Gaussian coordinates using the transformation function from the scikit-image library.
An example of georeferencing is shown in Figure~\ref{fig:gclf}\red{c}.

\noindent \textbf{Inference.}
Given a query $q$, GCLF localizes the target object by computing the cosine similarity between the CLIP text feature $T(q)$ and the decoded language embedding $D(\hat{\bm{l}}(v))$ at each pixel $v$ of a rendered 2D image
Here, $T$ is the CLIP text encoder, $D$ is the decoder of the autoencoder, and $\hat{\bm{l}}(v)$ is the aggregated language embedding at pixel $v$.
\if[]
To ensure the efficiency of this computation, GCLF dynamically leverages its hierarchical structure. 
When rendering each pixel, the system automatically selects the optimal hierarchical level where the projected diameter of the 3D Gaussians is smaller than one pixel. 
This strategy avoids computing similarities across all Gaussians in the entire scene, significantly reducing computational costs and enabling near real-time interaction.
\fi
To ensure efficiency, GCLF leverages its hierarchy to render each pixel using only Gaussians projected smaller than one pixel.
This avoids exhaustive similarity computations across the entire scene and enables real-time interaction.

\subsection{Geographical Vision APIs }
\label{sec:method:geo-vision-apis}

The functions of the 3D language fields, including GCLF, are primarily limited to the localization for objects corresponding to word-level natural language queries. 
To leverage GCLF for a variety of tasks,
GV-APIs provide a suite of visual and geographic processing functions.
Table~\ref{tab:gvapi_list} shows how each function is called and its role.
APIs \textbf{1)} to \textbf{6)} retrieve the relevant 3D Gaussians within GCLF, effectively narrowing down the region of interest in the vast city space (as shown in Figure~\ref{fig:qualitative___} GRD SEG2).
Specifically, \textbf{2)} selects 3D Gaussians that closely match the query based on cosine similarity (e.g., Figure~\ref{fig:qualitative___} GRD Answer). APIs \textbf{7)} and \textbf{8)} compute distances within GCLF, while \textbf{9)} applies GroundingDINO on RGB renderings from GCLF.
All these APIs work on trained GCLF without requiring additional training. Its technical contribution is to allow operations over the trained 3D Gaussian space.
For example, \textbf{8)} estimates landform height by identifying horizontal planes from Gaussian variance directions, while \textbf{6)} applies clustering to filter out noisy activations and improve segmentation accuracy.

\definecolor{codeboxcolor}{RGB}{4,132,0}
\definecolor{codeboxcolorl}{RGB}{239,255,238}
\newtcolorbox{codebox}[1]{colback=codeboxcolorl!30!white,colframe=codeboxcolor,fonttitle=\bfseries,title=#1,left=0mm,right=0mm,top=0mm,bottom=0mm,toptitle=-0.2mm,bottomtitle=-0.8mm}

\lstset{
 	language = sh,
 	breaklines = true,
 	breakindent = 0pt,
 	basicstyle = \ttfamily\scriptsize,
 	commentstyle = {\itshape \color[cmyk]{1,0.4,1,0}},
 	classoffset = 0,
 	keywordstyle = {\bfseries \color[cmyk]{0,0,0}},
 	stringstyle = {\ttfamily \color[rgb]{0,0,0}},
 	framesep = 5pt,
 	numbers = none,
 	stepnumber = 1,
 	numberstyle = \tiny,
 	tabsize = 4,
 	captionpos = t
}

\begin{figure}
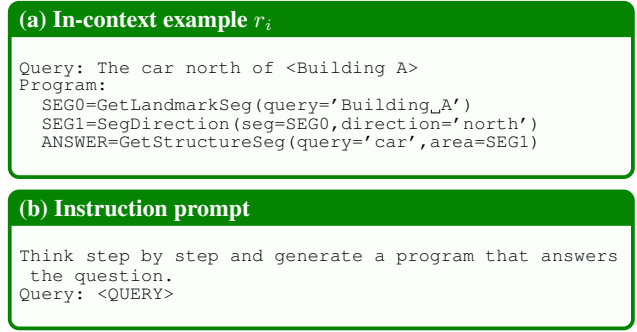

\vspace{-0.5em}

\centering
\begin{codebox}{\small (a) In-context example $r_{i}$}
\begingroup
\renewcommand{\baselinestretch}{0.7}
\begin{lstlisting}
Query: The car north of <Building A>
Program:
  SEG0=GetLandmarkSeg(query='Building A')
  SEG1=SegDirection(seg=SEG0,direction='north')
  ANSWER=GetStructureSeg(query='car',area=SEG1)
\end{lstlisting}
\endgroup
\end{codebox}
\vspace{-8pt}
\begin{codebox}{\small (b) Instruction prompt}
\begingroup
\renewcommand{\baselinestretch}{0.7}
\begin{lstlisting}
Think step by step and generate a program that answers the question.
Query: <QUERY>
\end{lstlisting}
\endgroup
\end{codebox}
\vspace{-1.3em}
\caption{ICE and instruction prompt.}
\label{fig:prompt}
\figvspace
\vspace{-0.7em}
\end{figure}

\begin{table*}[!t]
\vspace{-1.5em}
\centering
\setlength{\tabcolsep}{8.2pt}
\footnotesize
\begin{tabular}{llll}\toprule
&Function &Roles \\\midrule
1) &GetLandmarkSeg(query=QUERY) &Get segment by landmark name (\textit{e.g.}, ``Trinity Church''). \\
2) &GetStructureSeg(query=QUERY, area=SEG) &Get segment by structure name (\textit{e.g.}, ``bridge'' and ``tower''). \\
3) &SegAround(area=SEG, distance=DIST) &Get segment around the input segment according to DIST. \\
4) &SegDirection(area=SEG, direction=DIR) &Get segment with the specified direction for the input segment according to DIR. \\
5) &SegBetween(seg1=SEG1, seg2=SEG2) &Get segment located between two input segments. \\
6) &LargestSeg(segs=SEG) &Get the largest contiguous segment of the input segments using clustering. \\
7) &MeasureDist(from=SEG1, to=SEG2) &Get real-world distance (meters) between two input segments. \\
8) &MeasureHeight(area=SEG) &Get real-world height (meters) of input segment. \\
9) &GetObjectSeg(query=QUERY, area=SEG) &Execute GroundingDINO object detection by object name (\textit{e.g.}, ``car''). \\
\bottomrule
\end{tabular}
\vspace{-1em}
\caption{
How to call the nine functions of the GV-APIs and their roles.
}
\label{tab:gvapi_list}
\vspace{-1.5em}
\end{table*}

\subsection{Visual programming}
\label{sec:method:visprog}
To enable compositional reasoning through dynamic combinations of GV-APIs, we utilize visual programming.
Following previous studies on visual programming~\cite{Gupta2022VisProg, Suris2023ViperGPT}, an LLM $\Pi$ is utilized to generate a Python program $z$ given an input query $q$ with in-context examples (ICEs) $R$.
Specifiaclly, the GPT-3.5 model (\scalebox{0.95}[1]{\texttt{gpt-3.5-}}
\scalebox{0.95}[1]{\texttt{turbo-instruct}}) is used as $\Pi$. To instruct $\Pi$ on how to use GV-APIs $\mathcal{Z}$, we provide ten $R$. Each example consists of a short query paired with an example program.
Figure~\ref{fig:prompt} shows an example of $R$ and instruction prompts for the grounding task
(task details are provided in Section~\ref{sec:dataset:task}).
The generated program $z$ is executed by the Python execution engine $\Lambda$ with pre-trained GCLF $\mathcal{T}$ to obtain the answer $a$. $z$ is executed within a try block and returns None if an error is encountered during execution.
Figure~\ref{fig:exec} shows an example of the entire process of localizing a specific object (``Red-letter billboard'') by combining GV-APIs.
Three functions contribute to the compositional reasoning procedure.

Our framework requires only a small number of ICEs for the LLM to effectively utilize the GV-APIs. Empirically, providing 10-15 examples enables the LLM to generate programs for diverse queries with a high success rate of over 90\%. Crucially, the LLM does not merely imitate the provided examples. It combines its understanding of the API definitions with its pre-trained knowledge (e.g., concepts of counting and comparison) to generate programs with novel structures not seen in the examples, demonstrating strong structural-level generalization. This allows the framework to respond robustly even to variations in query phrasing (see Appendix~\ref{sec:ap_ice_count} for a detailed analysis).

\if[]
\noindent \textbf{Program generation.}
Following previous studies on visual programming~\cite{Gupta2022VisProg, Suris2023ViperGPT}, an LLM is utilized to generate a Python program given an input query with in-context examples (ICEs).
Specifiaclly, the GPT-3.5 model (\scalebox{0.95}[1]{\texttt{gpt-3.5-}}
\scalebox{0.95}[1]{\texttt{turbo-instruct}}) is used as an LLM. To instruct the LLM on how to use GV-APIs, we provide ten ICEs. Each example consists of a short query paired with an example program. Figure~\ref{fig:prompt} shows an example ICE for the grounding task and the instruction prompt (task details are provided in Section~\ref{sec:dataset:task}).

\noindent \textbf{Program execution.}
The generated program is executed by the Python execution engine with the pre-trained GCLF. The program is executed within a try block and returns None if an error is encountered during execution.
\fi

\begin{table} 
\centering
\setlength{\tabcolsep}{1pt}
\scriptsize
\begin{tabular}{l|p{8cm}}\toprule
Task & Query examples \\
\midrule
\multirow{3}{*}{GRD}& U-shaped building to the west of \textit{Liberty Luxe} \\
& Red canopy shop within 150 meters of \textit{Chase Bank}. \\
& Church with blue copper domes \\
      
\midrule
\multirow{3}{*}{CNT} & How many sports fields are there? \\
& How many ships are there northwest of \textit{200 Vesey Street}? \\
& How many cars are there between \textit{Little Stadium} and \textit{Huiyi building}? \\

\midrule
\multirow{3}{*}{MES}& How tall is the skyscraper near \textit{Quik Park}.    \\
& How tall is the building with a gray curved roof?\\
& How far is the fountain that is closest to \textit{Washington\hspace{1pt}Square\hspace{1pt}Arch} from \textit{Amity\hspace{1pt}Hall}?\\

\midrule
\multirow{3}{*}{CMP} & Which is taller, the cubic building or yellow building? \\
& Which is taller, \textit{The View} or \textit{Quik Park}? \\
& Which is taller, the tallest object around \textit{Quik Park} or \textit{The Avalon}? \\

\midrule
\multirow{3}{*}{SPR} & There is a light-blue pointed roof.   \\
& There is one or more tennis courts around \textit{Amity Hall}.\\
& The skyscraper that is closest to \textit{Quik Park} is closer to \textit{The View} than \textit{Quik Park}.\\      

\bottomrule
\end{tabular}
\vspace{-1em}
\caption{
Query examples for each task.
}
\label{tab:query_examples}
\vspace{-1em}
\end{table}



\section{GeoEval3D Dataset}
In this section, we present five tasks 
for evaluating understanding of city-scale 3D scenes, and introduce \datasetname, a dataset covering these tasks.
The dataset $\mathcal{B} = \{(\mathcal{D}_{i}, \mathcal{Q}_{i})\}_{i=1}^{S}$ consists of pairs multi-view image sets $\mathcal{D}_{i}$ and task sets $\mathcal{Q}_{i}$, where $S$ is the number of outdoor scenes.

\subsection{Task Definition}
\label{sec:dataset:task}
The task set $\mathcal{Q}_{i} = \{(q_{k}, a_{k})\}_{k=1}^{K_{i}}$ consists of pairs of queries $q_{k}$ and the corresponding ground truth answers $a_{k}$.
Each pair represents one of the five tasks listed in Table~\ref{tab:query_examples} with examples. The task definitions are summarized below. Further details are provided in Appendix B.

\noindent \textbf{1) Grounding (GRD).} Given a query $q_{k}$ describing a specific target object, this task requires models to identify and localize the object.
Following~\cite{kerr2023LERF,qin2023LangSplat}, the ground truth $a_{k}$ is the segment of the target object.

\noindent \textbf{2) Counting (CNT).} Given a query $q_{k}$ that involves counting objects in a scene, this task requires models to accurately count the number of specified objects.
The ground truth $a_{k}$ is provided as an integer.

\noindent \textbf{3) Measuring (MES).} Given a query $q_{k}$ describing a question, this task requires models to accurately measure the height (MES-H) and distance (MES-D) of buildings.
The ground truth $a_{k}$ is provided as an integer.

\noindent \textbf{4) Comparison (CMP).} 
Given a query $q_{k}$ that involves comparing the sizes of buildings, this task requires models to return a text $a_{k}$ identifying the correct building.

\noindent \textbf{5) Spatial reasoning (SPR).} Given a query $q_{k}$ describing 
object details or spatial relationships, this task requires models to return
``yes'' if it is correct, and ``no'' otherwise. 
The ground truth is provided as $a_{k} \in \{\text{yes}, \text{no}\}$.

\begin{figure}
\vspace{-0.5em}
\centering
\includegraphics[width=\linewidth]{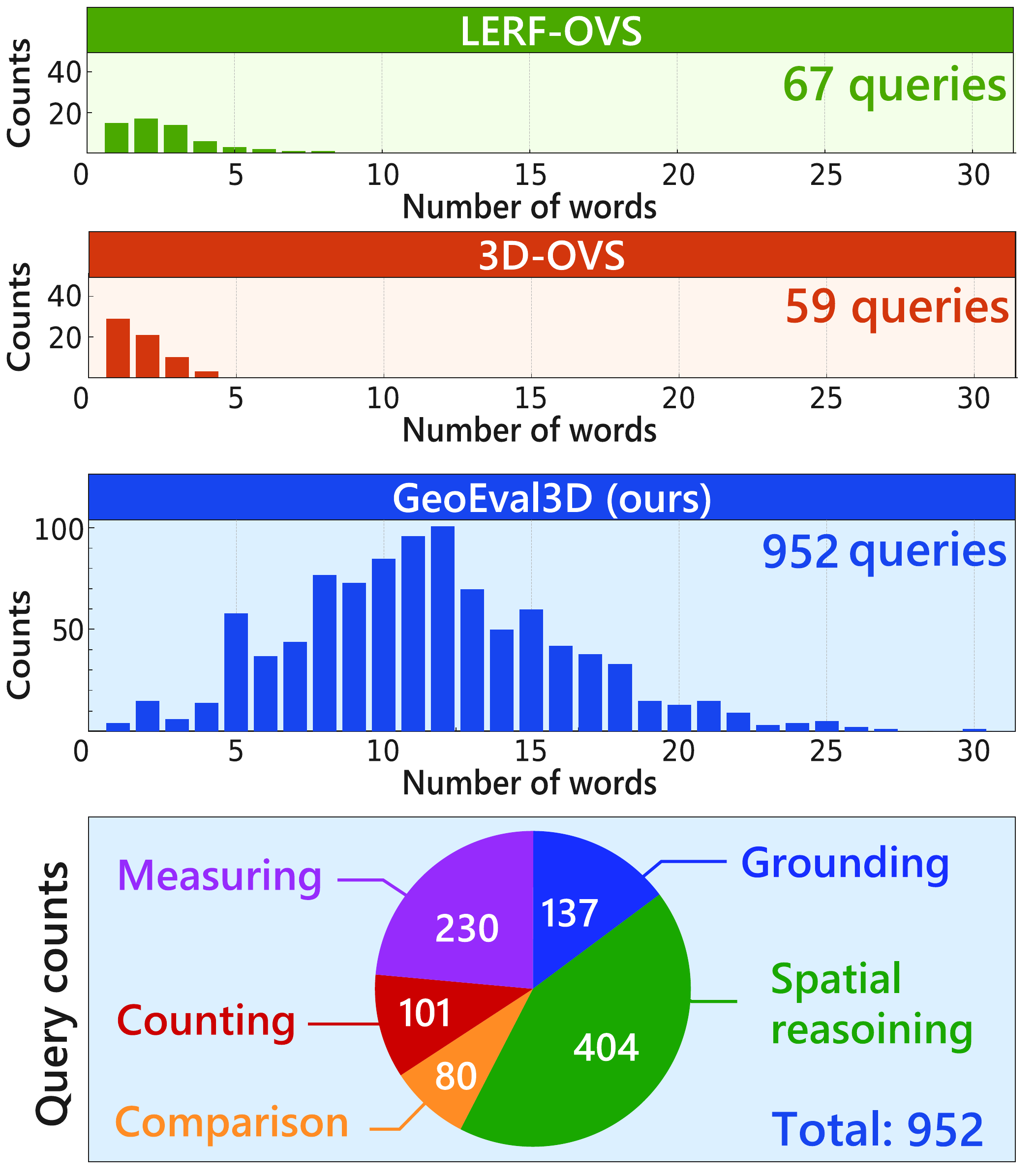}
\vspace{-2em}
\caption{Word count and query distributions.
GeoEval3D contains 952 unique queries covering five tasks.
Queries include more words than those in the previous evaluation datasets, indicating complexity of the proposed.}
\label{fig:annotation_stats}
\vspace{-1.5em}
\end{figure}

\begin{table*}[!t]
\vspace{-1em}
\centering
\setlength{\tabcolsep}{8.2pt}
\scriptsize
\begin{tabular}{lccccccccc}
\toprule
& \multicolumn{5}{c}{GoogleEarth} & \multicolumn{3}{c}{UrbanScene3D} \\
  \cmidrule(r){2-6} 
  \cmidrule(l){7-9} 
Method & \makecell{SPR \\ Acc.$\uparrow$} & \makecell{CMP \\ Acc.$\uparrow$} &  \makecell{CNT \\ MAE$\downarrow$} & \makecell{MES-H \\ MAE ($m$)$\downarrow$} & \makecell{MES-D \\ MAE ($m$)$\downarrow$} & \makecell{SPR \\ Acc.$\uparrow$} &  \makecell{CNT \\MAE $\downarrow$} & \makecell{MES-D \\ MAE ($m$)$\downarrow$} \\
\midrule
GPT-4o Vision~\cite{openai2024hello}                  & 24.77 &  2.63 & 3.02 & 158.16 & 195.29 & 15.18 & 4.29 & 1583.48&\\
LLaVA-1.5~\cite{liu2023llava}  & 50.95 & 36.96 & 3.08 & 607.37 & 433.15 & 46.04 & 4.23 & 837.11\\
Llama-3.2 Vision~\cite{meta2024llama32}    & 54.84 & 28.49 & 2.54 & 88.06 & 133.20 & 57.34 & 3.54 & 427.94&\\
Qwen2.5-VL-7B~\cite{Shuai2025QwenVL}        &53.39 &26.95 &2.65 &59.68 &175.44 &47.24 &4.00 &412.86 \\
InternVL2.5-8B~\cite{chen2024internvl}      &54.27 &26.95 &2.79 &51.30 &157.14 &52.47 &4.23 &318.71 \\
GeoChat~\cite{Kuckreja_2024_CVPR} & 57.23 & 41.99 & 2.89 & 84.74 & 89.34 & 56.76& 3.69 & 328.68& \\
LHRS-BOT~\cite{dilxat2024lhrs} & 49.45 & 27.52 & 4.85 & 46.17 & 104.49 & 41.94 & 3.49 & 438.94&\\
VHM~\cite{pang2024vhmversatilehonestvision} &54.55 &39.82 &5.28 &52.58 &135.50 &56.92 &4.34 &354.91 \\
TEOChat~\cite{irvin2024teochat} &59.04 &48.11 &2.84 &150.39 &198.89 &57.99 &4.06 &359.71 \\
\textbf{GeoProg3D} &\textbf{64.00} &\textbf{59.73} &\textbf{2.00} &\textbf{45.24} &\textbf{49.28} &\textbf{60.87} &\textbf{2.51} & \textbf{139.51}\\
\bottomrule
\end{tabular}
\vspace{-1em}
\caption{Performance of spatial reasoning (SPR), comparison (CMP), counting (CNT), and measurement (MES) tasks.}
\label{tab:compositional_reasoning}
\end{table*}

\subsection{Images}

GeoEval3D comprises urban scenes sourced from two datasets:
GoogleEarth~\cite{xie2024citydreamer} and the UrbanScene3D~\cite{Lin2022UrbanScene3D}.
Each scene has an image set $\mathcal{D}_{i} = \{x_{j}\}_{j=1}^{N}$ consists of multi-view images $x_{j}$ for training 3D scene representation.
From the GoogleEarth dataset, we chose four scenes, including scenes of 
New York (U.S.)
collected from Google Earth Studio.
Each scene comprises 60 images captured in an orbital pattern around a central subject. 
The orbit radius ranges from 125 meters to 813 meters, and the altitude varies from 112 meters to 884 meters. The images have a resolution of approximately 958 $\times$ 538 pixels, providing slightly coarse visuals suitable for large-scale urban modeling.
From the UrbanScene3D dataset, we chose one scene, a large real city in China (Shenzhen), covering a total area of 2 $km^2$. 
This scene includes multiple orbital images, offering high-definition visuals at approximately 4K resolution.
To enable the evaluation of various geographic vision tasks,
we provide annotations including object masks and distance measurements as ground truth for text queries.

\subsection{Dataset construction and statistics.}

\noindent \textbf{Annotation.}
All query-answer pairs were manually created by five annotators who were instructed to create high-quality ground truth.
To ensure the reliability of the dataset, annotation was performed using common tools.
Specifically, GT masks were created by a annotation tool of LabelMe.
Distance and height GTs were prepared by GoogleEarth, counting and yes/no correct labels were manually annotated through visual inspection.
Each query contains up to three landmark names.

\noindent \textbf{Statistics.} 
GeoEval3D is composed of unique 952 queries. As shown in Figure~\ref{fig:annotation_stats}, it is scaled up more than 10 times compared to the datasets used in previous works, and contains many more words~\cite{kerr2023LERF,qin2023LangSplat}.
The SPR task requires particularly complex compositional reasoning, thus accounting for about half of the total queries.

\begin{table}[t]
\centering
\setlength{\tabcolsep}{4pt}
\scriptsize
\begin{tabular}{l|c|ccccc}\toprule
Test scene & Area ($m^2$) & LSeg & LERF & LangSplat & GCLF & GeoProg3D \\
\midrule
GoogleEarth    & $2.4\times10^5$ & 0.96 & 11.44 & 14.15 & 20.09 & \textbf{45.20} \\
UrbanScene3D    & $5.0\times10^6$ & 4.65 & OOM & OOM &6.98 & \textbf{30.23} \\
\bottomrule
\end{tabular}
\vspace{-1em}
\caption{Localization accuracy (\%) on the GRD task.}
\label{tab:grounding:loc}
\vspace{-1em}
\end{table}


\begin{table}[t]
\centering
\setlength{\tabcolsep}{4pt}
\scriptsize
\begin{tabular}{l|c|ccccc}\toprule
Test scene & Area ($m^2$) & LSeg & LERF & LangSplat & GCLF & GeoProg3D \\
\midrule
GoogleEarth       & $2.4\times10^5$ & 1.08 & 6.38 & 5.19 & 6.69 & \textbf{18.15} \\
UrbanScene3D       & $5.0\times10^6$ & 1.06 & OOM &  OOM  & 3.78 & \textbf{8.74} \\
\bottomrule
\end{tabular}
\vspace{-1em}
\caption{3D semantic segmentation performance on the GRD task. Average IoU scores (\%) are reported.}
\label{tab:grounding:seg}
\vspace{-1em}
\end{table}



\begin{figure*}
\vspace{-1em}
\centering
\includegraphics[width=\linewidth]{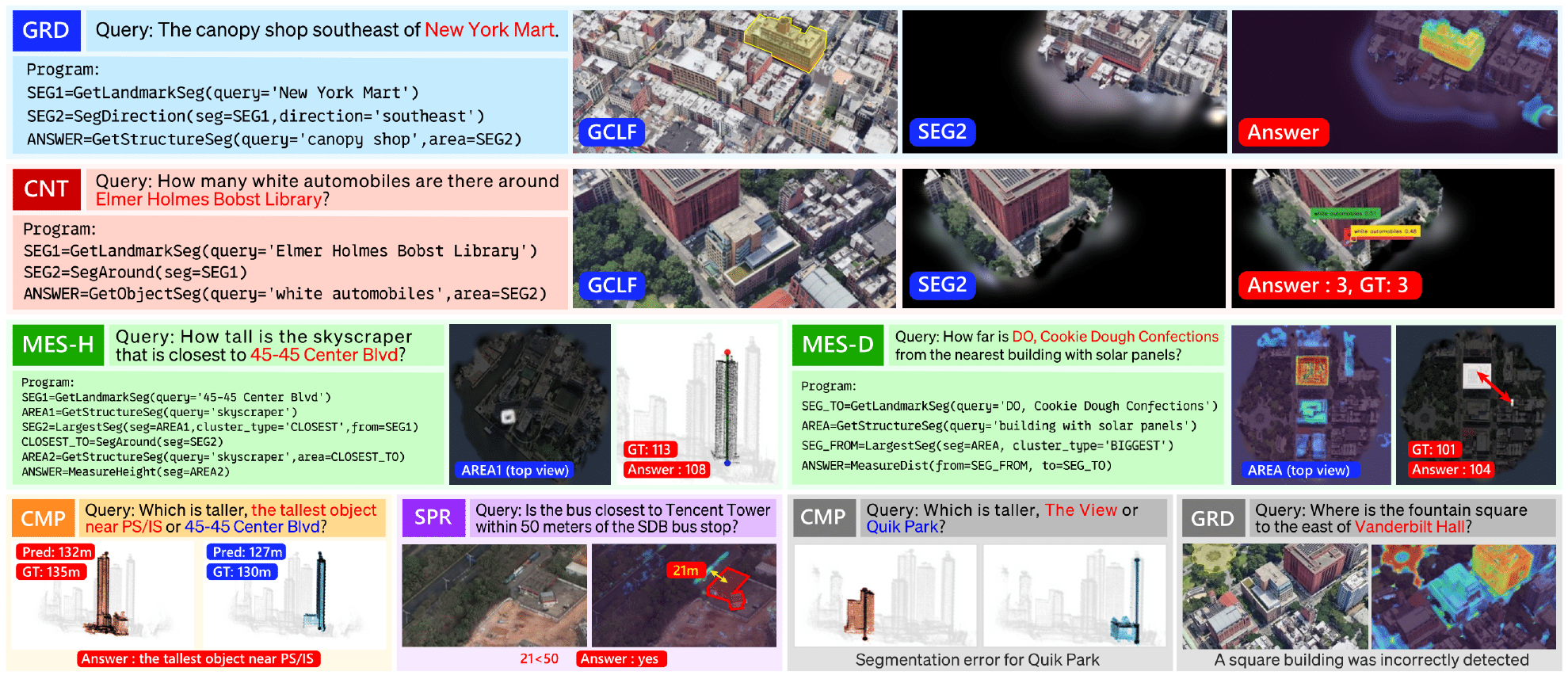}
\figvspacetop
\vspace{-1em}
\caption{Qualitative results and failure cases. The Ground Truth region for the GRD task is delineated by the yellow frame.}
\label{fig:qualitative___}
\vspace{-1.2em}
\end{figure*}


\section{Experiments}

\subsection{Evaluation metrics}
For the GRD task, we report localization accuracy for object localization and Intersection over Union (IoU) for segmentation following the previous study~\cite{qin2023LangSplat}.
Localization accuracy is measured at an IoU threshold of 0.15.
For the CNT and MES tasks, we calculate the Mean Absolute Error (MAE) between the predicted and true values. 
For the SPR and CMP tasks, exact match criteria are applied to determine correctness to compute accuracy.
We perform experiments on five scenes across the two datasets: four scenes from GoogleEarth and one scene from UrbanScene3D.
Note that MES-H and CMP are not evaluated in UrbanScene3D because Ground Truth for height cannot be obtained.

\subsection{Experimental results}
\noindent \textbf{Localization performance.}
In this experiment, we evaluate the localization performance of the 3D language field alone (GCLF) and when using GV-APIs through visual programming (GeoProg3D) in the GRD task.
We compare our methods with baselines, including LangSplat~\cite{qin2023LangSplat}, which is the SOTA method for high-resolution 3D scene localization.
The localization accuracies are shown in Table~\ref{tab:grounding:loc}.
We observed that GCLF outperforms baselines on GoogleEarth.
This suggests not only that language embedding into the tree structure works correctly, but also that high-fidelity reconstruction with the tree structure may help in more accurate localization at the pixel level.
In addition, LangSplat caused a memory error with UrbanScene3D in our setting, which implies the efficiency of the tree structure for learning larger scenes~\cite{liu2024citygaussian}.
GeoProg3D further improved accuracy on both GoolgeEarth and UrbanScene3D.
In terms of 3D semantic segmentation, we observed a similar trend in Table~\ref{tab:grounding:seg}.
It is worth noting that GeoProg3D exhibited superior performance on UrbanScene3D, which spans more than $2km^2$.
These results demonstrate the limitations of localization using 3D language fields alone in 3D urban scenes and the effectiveness of GV-APIs and visual programming in 
improving compositional reasoning.


\noindent \textbf{Various geographic vision tasks performance.}
While GCLF is limited to localization, GeoProg3D supports a wide variety of tasks through visual programming. This experiment evaluates the versatility of GeoProg3D in CNT, MES, SPR, and CMP tasks.
As baselines, we selected five VLMs: GPT-4o Vision~\cite{openai2024hello}, LLaVA-1.5~\cite{liu2023llava} and Llama 3.2 Vision~\cite{meta2024llama32}, Qwen2.5-VL-7B~\cite{Shuai2025QwenVL}, InternVL2.5-8B~\cite{chen2024internvl}, as well as four VLGFMs: GeoChat~\cite{Kuckreja_2024_CVPR}, LHRS-BOT~\cite{dilxat2024lhrs}, VHM~\cite{pang2024vhmversatilehonestvision}, and TEOChat~\cite{irvin2024teochat}.
Since VLMs can only process 2D images, we fed top-down view images instead of 3D data. 
Table~\ref{tab:compositional_reasoning} summarizes the results.
First, in the CNT and MES tasks, which assess the model's ability to accurately count and measure, our method achieved the lowest MAE values.
Notably, our method excelled in the MES-D (distance measurement) task, demonstrating precise horizontal spatial assessment capabilities. 
These results underscore the superior performance of GeoProg3D in estimating quantities within large-scale 3D scenes and highlight the effectiveness of the program-based inference procedures.
In addition, in the SPR and CMP tasks, which require more complex compositional reasoning, our method outperformed all other methods.
These results indicate not only its proficiency in understanding spatial relationship 
but also its effectiveness in comparing and contrasting them.
Furthermore, GeoProg3D showed examples of successful inference with discrimination of structures that are difficult to see in the top-down view, such as the sides of buildings and billboards.
This shows that 3D language fields enable localization that is independent of the viewpoint, taking into account the characteristics of structures which cannot be seen from directly above.
See appendix C for details.
Note that these comparison VLN-based baselines do not support pixel-level inference, so the GRD task is not evaluated.


\noindent \textbf{Ablation study.}
To assess the impact of each component of GeoProg3D, we conducted an ablation study to investigate the three tasks of GoogleEarth's GRD, SPR, and CMP.
There are also tasks and modules that are not included in the comparison for execution reasons.
Specifically, the CNT and MES tasks are not included in the experiment because they cannot be evaluated by accuracy rate, and the dedicated modules 7), 8), and 9) are also not included in the comparison.
The module 2) that retrieves similar areas of the query is not ablated because the GRD cannot be executed without it.
The results in Table~\ref{tab:ablation} show the significance of each component in maximizing the model's performance.
Among the components, omitting the \texttt{GetLandmarkSeg} module led to a significant drop in performance on all tasks, with GRD plummeting to 6.26\%, SPR to 15.77\%, and CMP to 0.00\%. This shows the vital role of segmentation for landmark objects.
Similarly, omitting \texttt{SegDirection} resulted in a marked decrease in GRD performance to 26.01\%, indicating the importance of directional cues in grounding tasks.
The omission of \texttt{SegAround} impairs performance particularly in the SPR task.
This indicates the necessity of identifying objects in close proximity for accurate spatial reasoning.
The exclusion of \texttt{SegBetween} impacts all tasks, though less drastically than other components.
Lastly, omitting the \texttt{LargestSeg} module affected the CMP performance, reducing the score to 44.74.
This highlights the importance of identifying the largest segment area for precise comparisons.
See appendix B for more ablation studies.

\noindent \textbf{Qualitative results and failure cases.}
Figure~\ref{fig:qualitative___} shows qualitative examples and failure cases. 
As shown, our approach successfully identifies the specified region to produce answers required for each task. For the example of CNT, when counting automobiles, the specified area is obtained as \texttt{SEG2}, and then the object detection function works to produce the answer that matches the ground truth. 
However, there are still challenging failure cases to address, as shown in the right bottom of Figure~\ref{fig:qualitative___}.
In GRD, there is an over-activation error caused by the square building responding to the square, which means a plaza.
In CMP, segmentation errors occur because geographic information is also assigned to parts of adjacent buildings.
Possible solutions for future work include redesigning the embedded language features and designing georeferencing to have a margin.

\begin{table}[t]
\vspace{-0.5em}
\centering
\footnotesize
\begin{tabular}{lccc}
\toprule
Method & GRD & SPR & CMP \\
\midrule
GeoProg3D & 45.20 &64.00 &59.73 \\
\hspace{2mm}w/o \texttt{GetLandMarkSeg} & 6.26 &15.77 &0.00 \\
\hspace{2mm}w/o \texttt{SegDirection}   & 26.01 &48.95 &52.63 \\
\hspace{2mm}w/o \texttt{SegAround}  & 34.67 &36.76 &43.42 \\
\hspace{2mm}w/o \texttt{SegBetween} & 40.14 &58.55 &60.53 \\
\hspace{2mm}w/o \texttt{LargestSeg} & 45.20 &51.51 &44.74 \\
\bottomrule
\end{tabular}
\vspace{-1em}
\caption{Ablation study of different Geographical Vision APIs.}
\label{tab:ablation}
\vspace{-1.5em}
\end{table}


\section{Conclusion}
We introduced GeoProg3D, a novel visual programming framework that enables human-computer interaction with city-scale 3D scenes through natural language queries. We also provided GeoEval3D for benchmarking 3D scene understanding models. Our experiments demonstrated that GeoProg3D significantly improves accuracy across the five visual geographical tasks. We believe our approach and dataset have contributed to the advancement of research in 3D scene understanding and visual programming.

{
    \small
    \bibliographystyle{ieeenat_fullname}
    \bibliography{main}
}

\appendix
\clearpage



\section{Additional Analysis on Visual Programming} 
\label{sec:ap_visprog_analysis}
This section provides a detailed analysis of the visual programming component in GeoProg3D, focusing on the impact of in-context examples and the framework's generalization capabilities.

\subsection{Effect of In-Context Example Count}
\label{sec:ap_ice_count}
\begin{figure}[h]
\vspace{-0.8em}
\centering
\includegraphics[width=\linewidth]{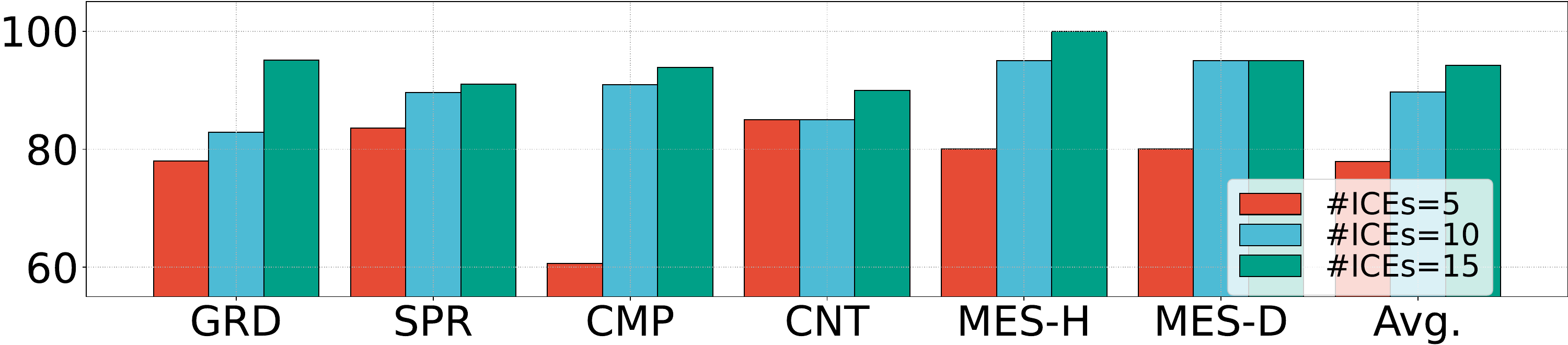}
\vspace{-1.2em}
\caption{
\small{Success rate of program generation for each task}
}
\label{fig:reb_nice}
\vspace{-1.2em}
\end{figure}

To evaluate the impact of the number of in-context examples (ICEs) on the performance of our framework, we conducted an experiment by varying the number of ICEs provided to LLM. As shown in Figure~\ref{fig:reb_nice}, we measured the success rate of program generation for each of our five tasks, using 5, 10, and 15 ICEs. The results demonstrate a clear trend: the program generation success rate improves as the number of ICEs increases. With 15 ICEs, the success rate reaches approximately 90\% for most tasks and begins to saturate. It is noteworthy that even with only 5 ICEs, GeoProg3D achieves a program generation success rate of around 70\%. This level of performance is sufficient to significantly outperform baseline methods. For instance, on the GRD task, GeoProg3D with 5 ICEs scores 38.02\% in localization accuracy, whereas LangSplat achieves only 17.07\%. This highlights the efficiency of our approach in leveraging LLMs for compositional reasoning with a minimal number of examples.

\subsection{Generalization to Rephrased Queries}

To assess the generalization capability and linguistic robustness of GeoProg3D, we evaluated its performance against rephrased queries. We used GPT-4o to generate paraphrased versions of the original queries in our GeoEval3D benchmark, creating a new test set denoted as "Rephrased Queries (RQ)" (Table~\ref{tab:reb_llm_rephrased_queries}). We then ran our full framework, GeoProg3D, on this RQ set. The results are presented in Table~\ref{tab:reb_robustness_grd} (for the GRD task) and Table~\ref{tab:reb_robustness_others} (for other tasks).

As the tables show, the performance of GeoProg3D with rephrased queries (GeoProg3D+RQ) is comparable to its performance with the original queries across all tasks. For instance, the SPR accuracy is 71.6\% for the original queries and 71.8\% for the rephrased ones, and the MAE for the MES-D task is 20.1m and 18.6m, respectively. This minimal difference in performance indicates that our framework is not reliant on specific keywords or phrasing. Instead, it demonstrates a strong ability to understand the semantic intent of a query and translate it into a correct executable program, highlighting the robust generalization capabilities of our approach.

\begin{table}[t]\centering
\scriptsize
\begin{tabular}{c|cccc}\toprule
Test scene & LangSplat &GCLF &GeoProg3D & GeoProg3D+RQ \\\midrule
GoogleEarth & 17.07 &26.83 &46.34 & 46.34 \\
\bottomrule
\end{tabular}
\caption{
\small{
Localization accuracy (\%) on the GRD task. 
}
}
\label{tab:reb_robustness_grd}
\vspace{-1em}
\end{table}

\begin{table}[t]\centering
\scriptsize
\begin{tabular}{lcccccc}\toprule
Method & \makecell{SPR \\ Acc.$\uparrow$} & \makecell{CMP \\ Acc.$\uparrow$} &  \makecell{CNT \\ MAE$\downarrow$} & \makecell{MES-H \\ MAE ($m$)$\downarrow$} & \makecell{MES-D \\ MAE ($m$)$\downarrow$} \\
\midrule
InternVL &64.2 &30.3 &1.8 &102.8 &115 \\
TEOChat &53.7 &63.6 &3.8 &236.2 &205.6 \\
GeoProg3D &71.6 &78.8 &1.4 &13.6 &20.1 \\
GeoProg3D+RQ &71.8 &78.8 &1.3 &13.2 &18.6 \\
\bottomrule
\end{tabular}
\caption{
\small{
Performance of SPR, CMP, CNT, and MES tasks.
}
}
\label{tab:reb_robustness_others}
\vspace{-2em}
\end{table}

\begin{table}[!htp]\centering
\vspace{-0.5em}
\scriptsize
\begin{tabular}{lll}\toprule
ID &Original queries \\\midrule
1 &How many buildings are there? \\
2 &There are at least two streets facing BldgA. \\
3 &There are 2 or less buildings to the directly west of BldgA. \\
4 &Which is taller, the tallest object around BldgA or BldgB? \\
5 &There are 2 or more grass areas to the directly west of BldgA. \\
\bottomrule
ID &Rephrased queries \\\midrule
1 &{What is the total number of buildings?} \\
2 &{BldgA faces at least two streets.} \\
3 &{At most two buildings stand directly to the west of BldgA.} \\
4 &{Which is taller, the tallest object surrounding BldgA or BldgB?} \\
5 &{To the direct west of BldgA, there are two or more grassy areas.} \\
\bottomrule
\end{tabular}
\caption{
Examples of test queries rephrased by LLM.
}
\label{tab:reb_llm_rephrased_queries}
\vspace{-1.0em}
\end{table}

\section{Implementation details} 
\label{sec:ap_imp_detail}

To extract object masks, the SAM model using the ViT-H backbone was  used.
For extracting CLIP features, the OpenCLIP ViT-B/16 model was used.
The tree-structure is implemented by utilizing the LoG rendering implementation~\cite{shuaiLoG2024}.
Each training stage is repeated 2,000, 15,000, and 30,000 times for the Google Earth dataset, and 300,000, 600,000, and 500,000 times for the UrbanScene3D dataset.
Our autoencoder is implemented using an MLP, which compresses 512-dimensional CLIP features into 3-dimensional latent features.
For GoogleEarch scenes, each composed of 60 images in with a resolution of 958$\times$538, training took about 15 minutes on an NVIDIA Quadro RTX 8000 GPT using 2GB of memory.
For the UrbanScene3D scene composed of 5,871 images with a resolution of 1620$\times$1080, training took about 6 hours using 40GB of memory. The total number of images used for training is 6,111.

\section{Technical contributions of GCLF} 
Our central contribution in designing GCLF lies in addressing the challenge of geometric distortion that arises from a naive integration of existing methods like LangSplat~\cite{qin2023LangSplat} and hierarchical representations~\cite{shuaiLoG2024}. 
To tackle this issue, GCLF employs a two-stage training strategy. 
First, it aligns the 3D Gaussians with real-world coordinates by referencing 2D geographic information from OpenStreetMap, and then freezes their geometric configuration. Subsequently, it learns hierarchical language features upon this fixed structure. This unique approach establishes GCLF as the first large-scale, hierarchical 3D language field for city-scale environments, achieving both high-fidelity scene representation and efficient language query processing.

\begin{table}[!t]
\centering
\setlength{\tabcolsep}{5pt}
\scriptsize
\begin{tabular}{l|c|ccccc}\toprule
Test Scene & Area ($m^2$) & LSeg &LERF & LangSplat & GCLF  & GeoProg3D \\
\midrule
Center Blvd     & $2.7\times10^5$ & 0.03 &0.07 &7.69 &19.23 &\textbf{42.31} \\
World Fin Ctr   & $4.7\times10^5$ & 0 &0.05 &20.00 &16.00 & \textbf{44.00} \\
Mott St         & $1.7\times10^5$ & 0 &0.05 &10.71 &17.86 & \textbf{53.57} \\
Washington Sq   & $1.3\times10^5$ & 0.01 &0.08 &18.18 &27.27 & \textbf{40.91} \\
Campus          & $5.0\times10^6$ & 0.01 &OOM &OOM &6.98 & \textbf{30.23} \\
\bottomrule
\end{tabular}
\caption{Localization accuracy (\%) for GRN task by scene.}
\label{tab:grounding_by_scene:loc}
\end{table}

\begin{table} [!t]
\centering
\setlength{\tabcolsep}{5pt}
\scriptsize
\setlength{\tabcolsep}{4.5pt}
\begin{tabular}{l|c|ccccc}\toprule
Test Scene & Area ($m^2$) & LSeg &LERF & LangSplat & GCLF  & GeoProg3D \\
\midrule
Center Blvd    & $2.7\times10^5$ &      0.04 &0.15 & 4.28 &6.10 & \textbf{19.74} \\
World Fin Ctr       & $4.7\times10^5$ & 0 &0.08 & 6.03 &6.31 & \textbf{16.12} \\
Mott St             & $1.7\times10^5$ & 0 &0.05 & 5.21 &7.56 & \textbf{25.60} \\
Washington Sq       & $1.3\times10^5$ & 0 &0.18 & 5.22 &6.80 & \textbf{11.12} \\
Campus              & $5.0\times10^6$ & 0.05 &OOM &OOM & 3.78 & \textbf{8.74} \\
\bottomrule
\end{tabular}
\caption{3D semantic segmentation performance for GNR task by scene. IoU scores (\%) are reporeted.}
\label{tab:grounding_by_scene:seg}
\end{table}

\begin{table*}[!t]
\centering
\setlength{\tabcolsep}{5pt}
\scriptsize
\begin{tabular}{lcccccccccc}\toprule
& \multicolumn{6}{c}{Spatial Reasoning: Accuracy (\%) $\uparrow$}  \\
\midrule
Test Scene & LLaVA-1.5 & Llama-3.2 Vision & GPT-4o Vision & Qwen2.5-VL & InternVL2.5 & GeoChat & TEOChat& LHRS-BOT & VHM & \textbf{GeoProg3D} \\
\midrule
Center Blvd &47.88 &49.23 &37.98 &56.86 &56.45 &58.41 &58.41 &49.43 &\textbf{63.47} &60.17 \\
World Fin Ctr &47.78 &62.65 &24.86 &52.21 &43.23 &61.67 &61.04 &53.19 &50.75 &\textbf{67.14} \\
Mott St &52.12 &53.87 &16.08 &50.73 &57.38 &58.41 &61.55 &50.37 &52.78 &\textbf{67.18} \\
Washington Sq &56.02 &53.62 &20.17 &53.74 &60.01 &50.41 &55.17 &44.83 &51.18 &\textbf{61.52} \\
Campus &46.04 &57.34 &15.18 &47.24 &52.47 &56.76 &57.99 &41.94 &56.92 &\textbf{60.87} \\
\bottomrule
\end{tabular}
\caption{SPR performance by scene. 
LLaVA-1.5~\cite{liu2023llava}, Llama-3.2 Vision~\cite{meta2024llama32}, GPT-4o Vision~\cite{openai2024hello}, Qwen2.5-VL~\cite{Shuai2025QwenVL}, InternVL2.5~\cite{chen2024internvl}, GeoChat~\cite{Kuckreja_2024_CVPR}, TEOChat~\cite{irvin2024teochat}, LHRS-BOT~\cite{dilxat2024lhrs}, VHM~\cite{pang2024vhmversatilehonestvision} and GeoProg3D are evaluated.}
\label{tab:spr_by_scene}
\end{table*}

\begin{table*}[!t]
\centering
\setlength{\tabcolsep}{5pt}
\scriptsize
\begin{tabular}{lcccccccccc}\toprule
& \multicolumn{6}{c}{Comparison: Accuracy (\%) $\uparrow$}  \\
\midrule
Test Scene & LLaVA-1.5 & Llama-3.2 Vision & GPT-4o Vision & Qwen2.5-VL & InternVL2.5 & GeoChat & TEOChat& LHRS-BOT & VHM & \textbf{GeoProg3D} \\
\midrule
Center Blvd &21.05 &\textbf{73.68} &0.00 &26.32 &57.89 &26.32 &57.89 &5.26 &36.84 &63.16 \\
World Fin Ctr &42.11 &21.05 &10.53 &52.63 &31.58 &52.63 &31.58 &36.84 &36.84 &\textbf{68.42} \\
Mott St &36.84 &10.53 &0.00 &15.79 &36.84 &36.84 &\textbf{42.11} &15.79 &\textbf{42.11} &\textbf{42.11} \\
Washington Sq &47.83 &8.70 &0.00 &13.04 &47.83 &52.17 &60.87 &52.17 &43.48 &\textbf{65.22} \\
\bottomrule
\end{tabular}
\caption{CMP performance by scene.
LLaVA-1.5~\cite{liu2023llava}, Llama-3.2 Vision~\cite{meta2024llama32}, GPT-4o Vision~\cite{openai2024hello}, Qwen2.5-VL~\cite{Shuai2025QwenVL}, InternVL2.5~\cite{chen2024internvl}, GeoChat~\cite{Kuckreja_2024_CVPR}, TEOChat~\cite{irvin2024teochat}, LHRS-BOT~\cite{dilxat2024lhrs}, VHM~\cite{pang2024vhmversatilehonestvision} and GeoProg3D are evaluated.}
\label{tab:cmp_by_scene}

\hspace{14pt}
\centering
\setlength{\tabcolsep}{5pt}
\scriptsize
\begin{tabular}{lcccccccccc}\toprule
& \multicolumn{6}{c}{Counting: MAE $\downarrow$}  \\
\midrule
Test Scene & LLaVA-1.5 & Llama-3.2 Vision & GPT-4o Vision & Qwen2.5-VL & InternVL2.5 & GeoChat & TEOChat& LHRS-BOT & VHM & \textbf{GeoProg3D} \\
\midrule
Center Blvd &2.23 &1.77 &1.77 &2.00 &1.85 &3.15 &2.46 &\textbf{1.54} &4.62 &1.92 \\
World Fin Ctr &2.86 &2.86 &3.07 &2.14 &3.00 &2.71 &2.93 &2.71 &2.21 &\textbf{1.93} \\
Mott St &3.84 &2.83 &3.68 &3.58 &3.42 &2.95 &2.89 &10.63 &7.68 &\textbf{1.63} \\
Washington Sq &3.39 &2.70 &3.57 &2.87 &2.91 &2.74 &3.09 &4.52 &6.61 &\textbf{2.52} \\
Campus &4.23 &3.54 &4.29 &4.00 &4.23 &3.69 &4.06 &3.49 &4.34 &\textbf{2.51} \\
\bottomrule
\end{tabular}
\caption{CNT performance by scene.
LLaVA-1.5~\cite{liu2023llava}, Llama-3.2 Vision~\cite{meta2024llama32}, GPT-4o Vision~\cite{openai2024hello}, Qwen2.5-VL~\cite{Shuai2025QwenVL}, InternVL2.5~\cite{chen2024internvl}, GeoChat~\cite{Kuckreja_2024_CVPR}, TEOChat~\cite{irvin2024teochat}, LHRS-BOT~\cite{dilxat2024lhrs}, VHM~\cite{pang2024vhmversatilehonestvision} and GeoProg3D are evaluated.}
\label{tab:cnt_by_scene}
\hspace{14pt}
\end{table*}

\begin{table*}[!t]
\vspace{-1em}
\centering
\setlength{\tabcolsep}{5pt}
\scriptsize
\begin{tabular}{lcccccccccc}\toprule
& \multicolumn{6}{c}{Measurement (Height): MAE (m$\downarrow$)}  \\
\midrule
Test Scene & LLaVA-1.5 & Llama-3.2 Vision & GPT-4o Vision & Qwen2.5-VL & InternVL2.5 & GeoChat & TEOChat& LHRS-BOT & VHM & \textbf{GeoProg3D} \\
\midrule
Center Blvd &418.37 &85.18 &231.21 &71.53 &57.79 &124.05 &193.58 &61.05 &53.00 &\textbf{50.99} \\
World Fin Ctr &349.89 &104.78 &198.26 &90.58 &91.26 &53.68 &184.32 &68.21 &59.26 &\textbf{58.16} \\
Mott St &962.12 &43.35 &167.73 &37.88 &\textbf{18.58} &62.12 &107.62 &22.65 &69.27 &56.28 \\
Washington Sq &699.09 &118.91 &35.43 &38.74 &37.57 &99.09 &116.04 &32.78 &28.78 &\textbf{15.52} \\

\bottomrule
\end{tabular}
\vspace{-1em}
\caption{MES-H performance by scene.
LLaVA-1.5~\cite{liu2023llava}, Llama-3.2 Vision~\cite{meta2024llama32}, GPT-4o Vision~\cite{openai2024hello}, Qwen2.5-VL~\cite{Shuai2025QwenVL}, InternVL2.5~\cite{chen2024internvl}, GeoChat~\cite{Kuckreja_2024_CVPR}, TEOChat~\cite{irvin2024teochat}, LHRS-BOT~\cite{dilxat2024lhrs}, VHM~\cite{pang2024vhmversatilehonestvision} and GeoProg3D are evaluated.}
\label{tab:mes-h_by_scene}
\vspace{-1em}
\end{table*}

\begin{table*}[!t]
\centering
\setlength{\tabcolsep}{5pt}
\scriptsize
\begin{tabular}{lcccccccccc}\toprule
& \multicolumn{6}{c}{Measurement (Distance): MAE (m$\downarrow$)}  \\
\midrule
Test Scene & LLaVA-1.5 & Llama-3.2 Vision & GPT-4o Vision & Qwen2.5-VL & InternVL2.5 & GeoChat & TEOChat& LHRS-BOT & VHM & \textbf{GeoProg3D} \\
\midrule
Center Blvd &169.95 &199.07 &196.84 &164.47 &165.42 &82.58 &164.42 &160.16 &141.05 &\textbf{61.60} \\
World Fin Ctr &460.21 &220.00 &285.05 &219.79 &239.79 &129.05 &178.47 &212.11 &162.47 &\textbf{67.94} \\
Mott St &306.58 &162.47 &164.68 &162.47 &125.21 &103.32 &264.05 &160.37 &142.47 &\textbf{55.00} \\
Washington Sq &98.95 &123.81 &283.10 &155.05 &98.14 &60.86 &188.62 &118.57 &96.00 &\textbf{28.71} \\
Campus &837.11 &427.94 &1583.48 &412.86 &318.71 &328.68 &359.71 &438.94 &354.91 &\textbf{139.51} \\
\bottomrule
\end{tabular}
\vspace{-1em}
\caption{MES-D performance by scene.
LLaVA-1.5~\cite{liu2023llava}, Llama-3.2 Vision~\cite{meta2024llama32}, GPT-4o Vision~\cite{openai2024hello}, Qwen2.5-VL~\cite{Shuai2025QwenVL}, InternVL2.5~\cite{chen2024internvl}, GeoChat~\cite{Kuckreja_2024_CVPR}, TEOChat~\cite{irvin2024teochat}, LHRS-BOT~\cite{dilxat2024lhrs}, VHM~\cite{pang2024vhmversatilehonestvision} and GeoProg3D are evaluated.}
\label{tab:mes-d_by_scene}
\vspace{-1em}
\end{table*}

\section{Detailed results and analysis} 
\label{sec:ap_detail}
\noindent \textbf{GRD task.}
Tables~\ref{tab:grounding_by_scene:loc} and ~\ref{tab:grounding_by_scene:seg} present the grounding performance by scene.
In all scenes, GeoProg3D achieves significantly higher performance compared to GCLF and other baselines.
Additionally, GCLF outperforms LangSplat in most scenes in terms of segmentation performance, indicating that its high-quality 3D representation leads to improved performance in the GRD task.


\vspace{2pt}
\noindent \textbf{SPR task.}
Table~\ref{tab:spr_by_scene} presents the spatial reasoning performance by scene, demonstrating that GeoProg3D
achieves the highest or competitive accuracy in all scenes.
Its accuracy remains above 60\% in every scene, showcasing strong generalization across diverse environments. 
The other models including GeoChat and Llama-3.2 Vision perform moderately well but lag behind, while GPT-4o Vision struggles with accuracies below 40\% in most cases. 

\vspace{2pt}
\noindent \textbf{CMP task.}
Table ~\ref{tab:cmp_by_scene} shows that GeoProg3D outperforms
the other models in three out of four scenes for the CMP task.
While Llama-3.2 Vision shows strong performance in Center Blvd (73.68\%), its accuracy drops significantly in the other scenes.
InternVL2.5 and VHM perform moderately well but fall short of GeoProg3D. LLaVA-1.5 and GPT-4o Vision struggle significantly, with the latter showing near-zero performance across most scenes. 

\begin{figure*}
\vspace{-1em}
\centering
\includegraphics[width=\linewidth]{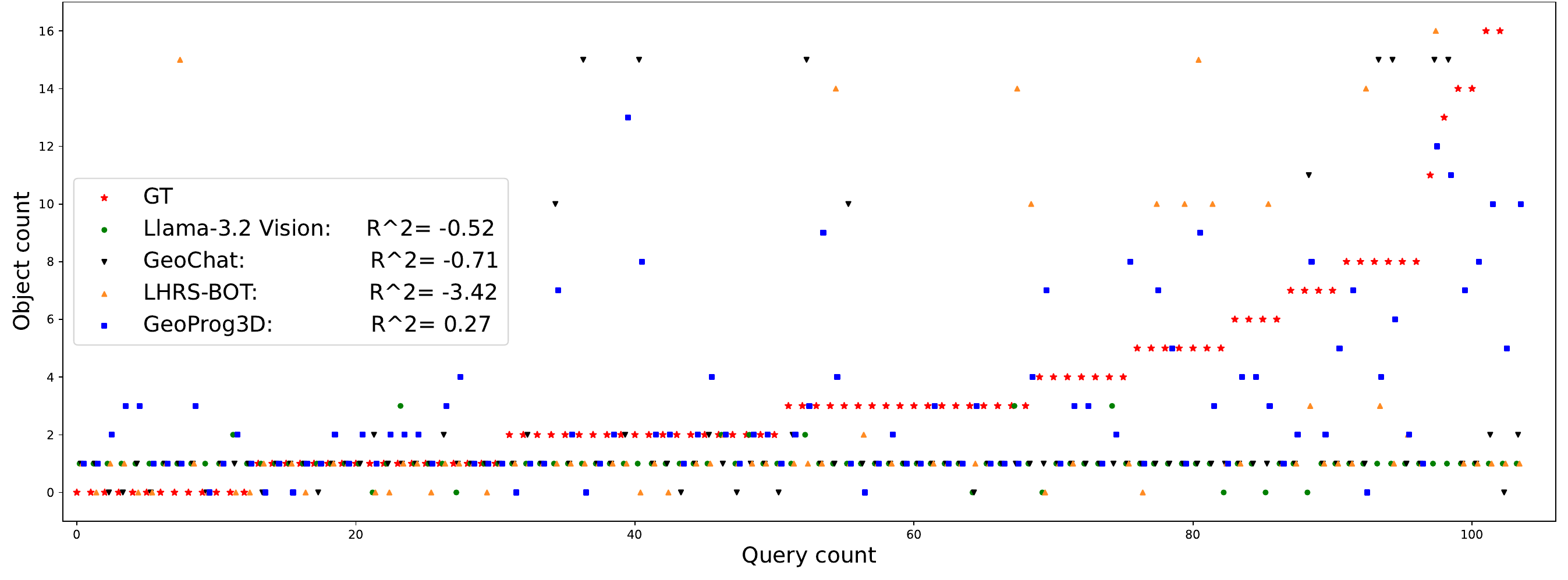}
\figvspacetop
\vspace{-1em}
\captionof{figure}{Comparison of predicted counts versus ground truth across different methods.}
\label{fig:cnt_analysis}
\figvspace
\end{figure*}

\vspace{2pt}
\noindent \textbf{CNT task.}
As shown in Table~\ref{tab:cnt_by_scene}, GeoProg3D also demonstrates strong performance in the CNT task, achieving competitive Mean Absolute Error (MAE) values across all scenes.
While LHRS-BOT achieves slightly better MAE in Center Blvd (1.54), GeoProg3D consistently performs well, maintaining low MAEs across all scenes.
LLaVA-1.5 and GPT-4o Vision show higher errors, indicating limited reliability in this task. 
Qwen2.5-VL and Llama-3.2 Vision perform moderately but lack the consistency shown by GeoProg3D. 
Table~\ref{tab:cnt_gt_average} and Figure~\ref{fig:cnt_analysis} show the average of ground truth counts by scene and a comparison of predicted and ground-truth counts, respectively.
The R-squared values in Figure~\ref{fig:cnt_analysis} indicate that methods other than GeoProg3D poorly align with the ground truth distribution. 
These methods often output ``1'' as the answer to the query, which may explain the smaller MAE observed in some cases for the Center Blvd scene.
These results highlight GeoProg3D's capability to balance precision and generalization in diverse counting scenarios.

\begin{figure*}
\centering
\includegraphics[width=\linewidth]{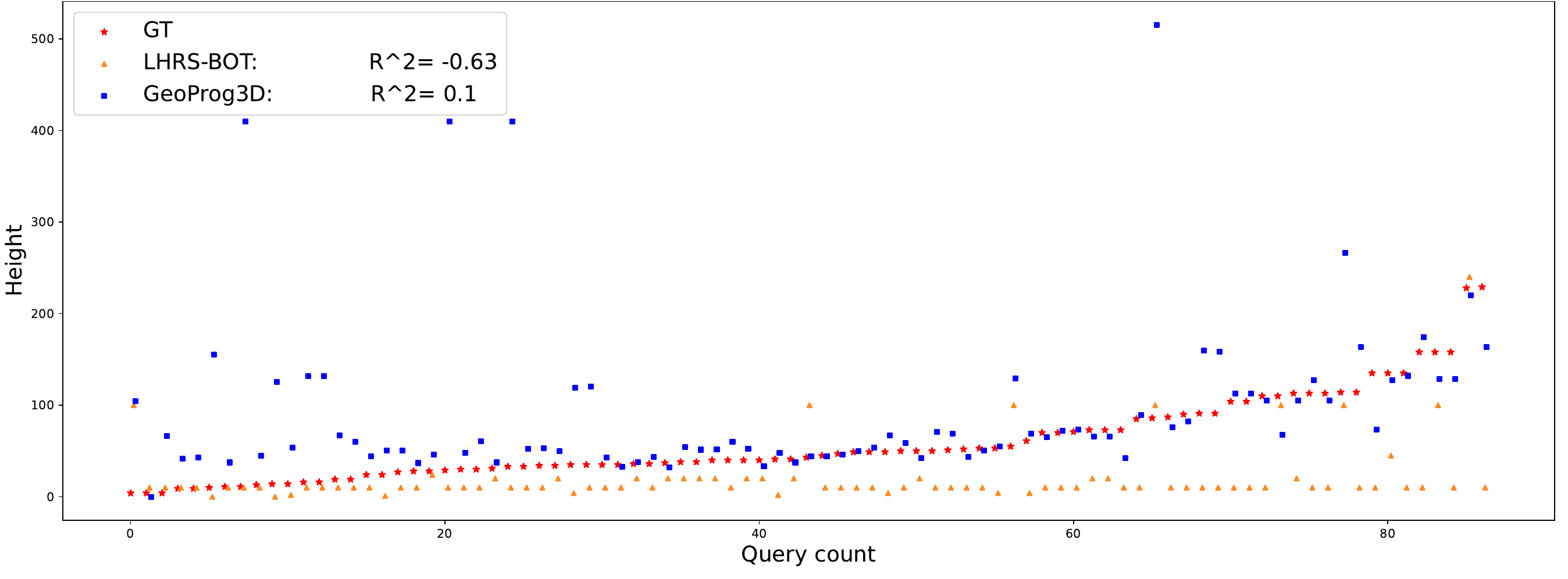}
\figvspacetop
\vspace{-1em}
\caption{Comparison of predicted height versus ground truth across different methods.}
\label{fig:mes-h_analysis}
\figvspace
\vspace{4pt}
\end{figure*}
\begin{figure*}[!t]
\centering
\includegraphics[width=\linewidth]{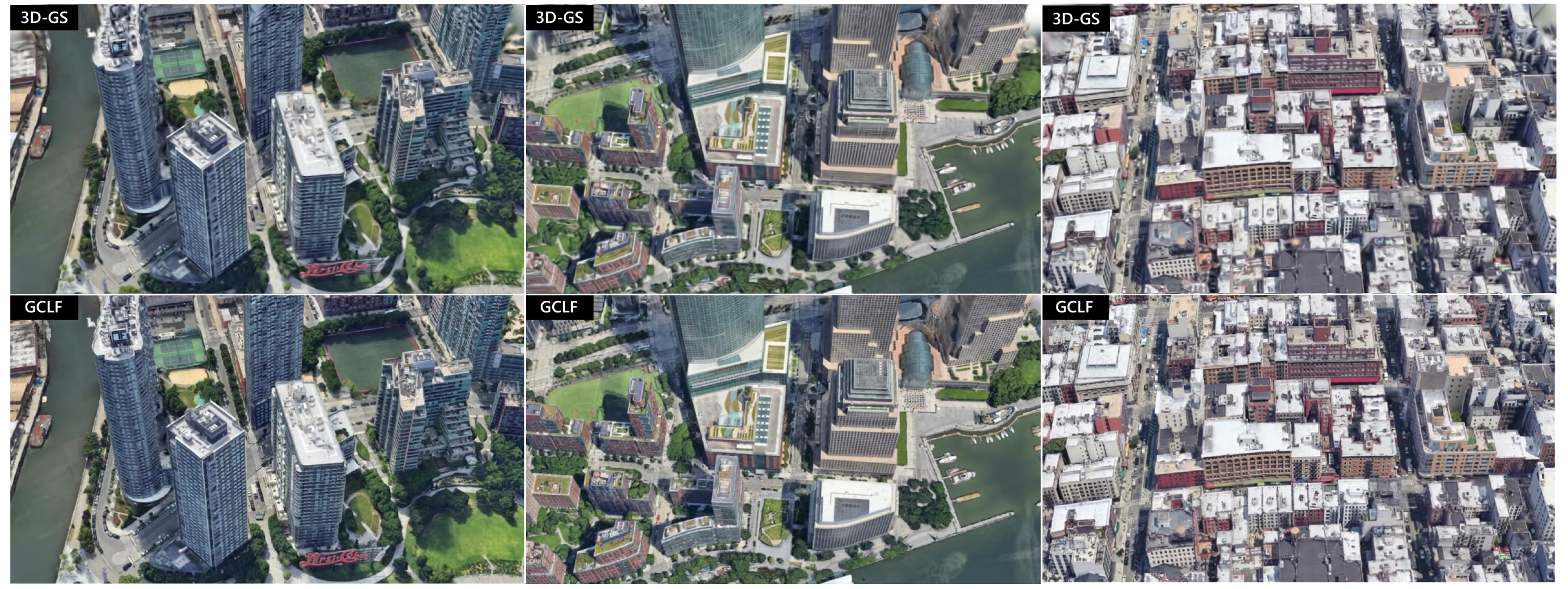}
\figvspacetop
\vspace{-1em}
\caption{Comparison of 3D scene reconstruction quality between 3D-GS and GCLF.}
\label{fig:recon_quality}
\figvspace
\end{figure*}

\vspace{2pt}
\noindent \textbf{MES task.}
Table~\ref{tab:mes-h_by_scene} shows that GeoProg3D demonstrates superior performance in both height (MES-H) and distance (MES-D) measurement tasks across most test scenes, as indicated by its consistently low MAE values. 
For MES-H, GeoProg3D outperforms the other methods in the scenes except for Mott St.
In the Mott St scene, there are 26 queries for MES-H, and GeoProg3D was unable to provide answers for 3 of them.
This issue is caused by an error in the program generation by the LLM.
Additionally, the Mott St scene contains many low-lying buildings (as shown in the rightmost scene in Figure~\ref{fig:recon_quality}), where baseline method achieves low error by consistently providing monotonous responses with small values, as illustrated in Figure~\ref{fig:mes-h_analysis}.
For MES-D, GeoProg3D achieves the lowest MAE across all test scenes, highlighting its precision in distance measurement. 
Notably, while most methods struggle with higher errors in a large environment like Campus, where GPT-4o Vision has an MAE exceeding 1500 meters, \methodname showed superior performance on this scene.


\begin{table} 
\centering
\small
\begin{tabular}{lrrrr}
\toprule
LLM &GRD &SPR &CMP \\\midrule
GPT-3.5 &45.20 &64.00 &59.73 \\
GPT-4o &46.09 &64.40 &64.76 \\
\bottomrule
\end{tabular}
\caption{Ablation of \methodname using different LLM backbones.}
\label{tab:llm_ablation}
\end{table}

\vspace{2pt}
\noindent \textbf{Ablation study using different LLMs.}
Table~\ref{tab:llm_ablation} compares the performance of GeoProg3D using different LLM backbones, specifically GPT-3.5 and GPT-4o, across three tasks: GRD, SPR, and CMP.
The results indicate that GPT-4o consistently outperforms GPT-3.5 across all tasks, with notable improvements in 
CMP (64.76\% compared to 59.7\%). 
The differences in GRD and SPR are smaller.
The improvements in 
CMP are particularly significant, indicating GPT-4o’s ability to handle more complex tasks. 
This highlights the potential to enhance GeoProg3D’s performance across a wide range of metrics by utilizing an advanced LLM backbone, depending on the available budget.


\begin{table} 
\centering
\setlength{\tabcolsep}{5pt}
\small
\begin{tabular}{lrrrr}\toprule
Scene &Avg. CNT &Avg. MES-H &Avg. MES-D \\\midrule
Center Blvd &2.31 &75.21 &161.21 \\
World Fin Ctr &3.07 &96.74 &220.00 \\
Mott St &3.74 &37.88 &162.47 \\
Washington Sq &3.61 &40.04 &123.81 \\
Campus &4.31 &- &427.94 \\
\midrule
Overall &3.63 &59.46 &284.94 \\
\bottomrule
\end{tabular}
\caption{Average ground truth values for CNT, MES-H, and MES-D queries.}
\label{tab:cnt_gt_average}
\end{table}

\noindent \textbf{Reconstruction quality.}
Figure ~\ref{fig:recon_quality} compares the image reconstruction quality of GCLF and a vanilla 3D-GS in several scenes.
As a result of training with the same number of epochs, the vanilla 3D-GS lacks texture details.


\noindent \textbf{Statistics.}
Table~\ref{tab:comparison_3dgs_log_stats} presents statistics for LangSplat and GCLF in terms of the number of Gaussians and inference speed across various scenes. 
GCLF consistently generates significantly more number of Gaussians than LangSplat, achieving more detailed representations. However, this reduces the rendering speed, as LangSplat is consistently faster across all cases.
These results highlight the trade-off between achieving detailed scene representations and maintaining computational efficiency. 
Nevertheless, GCLF demonstrates practical usability, as they are capable of querying even large-scale data within realistic time frames, ensuring their suitability for real-world applications.



\begin{figure*}[!t]
\vspace{-1em}
\centering
\includegraphics[width=\linewidth]{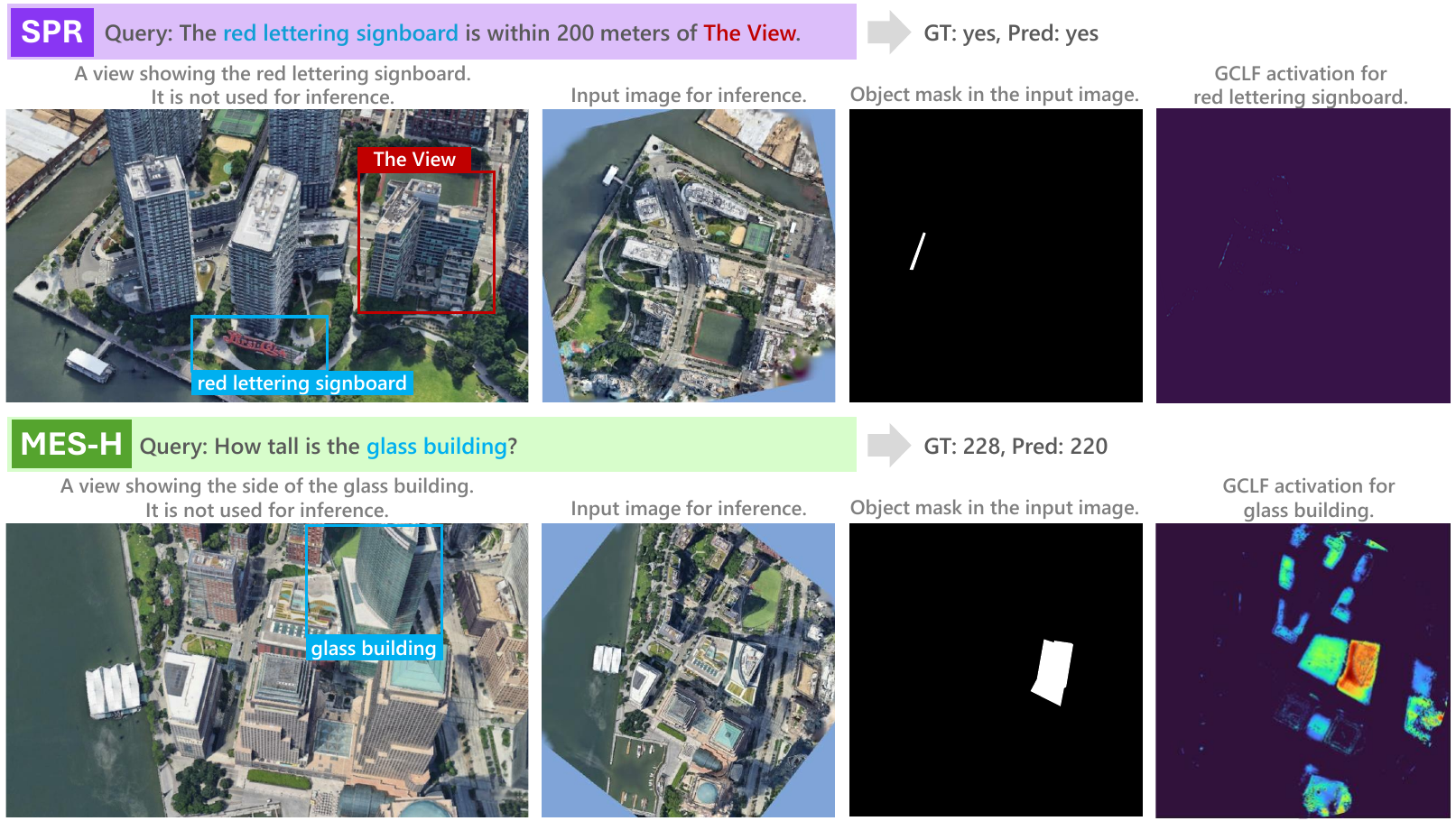}
\vspace{-2em}
\caption{
Examples of viewpoint-independent localization by GCLF.
}
\label{fig:3d_effectiveness}
\figvspace
\end{figure*}

\section{Qualitative examples}
\label{sec:ap_qualitative}

\begin{figure*}
\centering
\includegraphics[width=\linewidth]{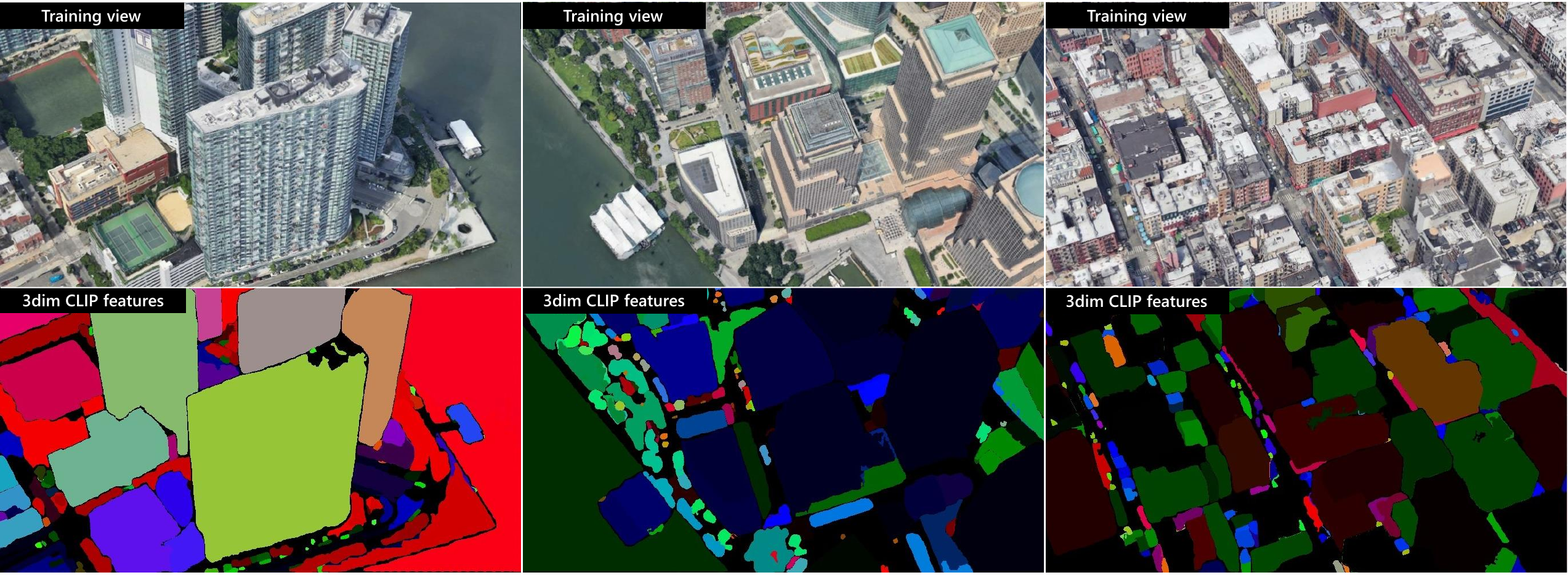}
\figvspacetop
\caption{Visualization of CLIP features for training 3D language fields.}
\label{fig:lang_feature}
\figvspace
\vspace{4pt}
\end{figure*}

\noindent \textbf{Visualization of CLIP features.}
Figure~\ref{fig:lang_feature} visualizes CLIP features projected into a three-dimensional feature space using an autoencoder during the training of the GCLF.
As illustrated, buildings and their surroundings are grouped into distinct clusters.

\noindent \textbf{Viewpoint-independent localization by GCLF.}
Since the VLMs that are compared in the versatility experiment can only process 2D images, all methods, including GeoProg3D, are input with top-down view images.
Information from directly above is not sufficient for searches related to the sides of buildings and signboards.
However, GCLF boosts the high performance of GeoProg3D by localizing it to take into account the characteristics of structures that cannot be seen from directly above.
Figure~\ref{fig:3d_effectiveness} shows examples of SPR and MES-H that have succeeded in reasoning in GeoProg3D with viewpoint-independent localization.
In SPR, the red lettering signboard is not visible in the top-down view used for inference, but it is correctly activated in the localization of the inference process.
In MES-H, whether a building is glass-fronted or not cannot be seen in the top-down view, but localization succeeded.

\begin{figure*}
\centering
\includegraphics[width=\linewidth]{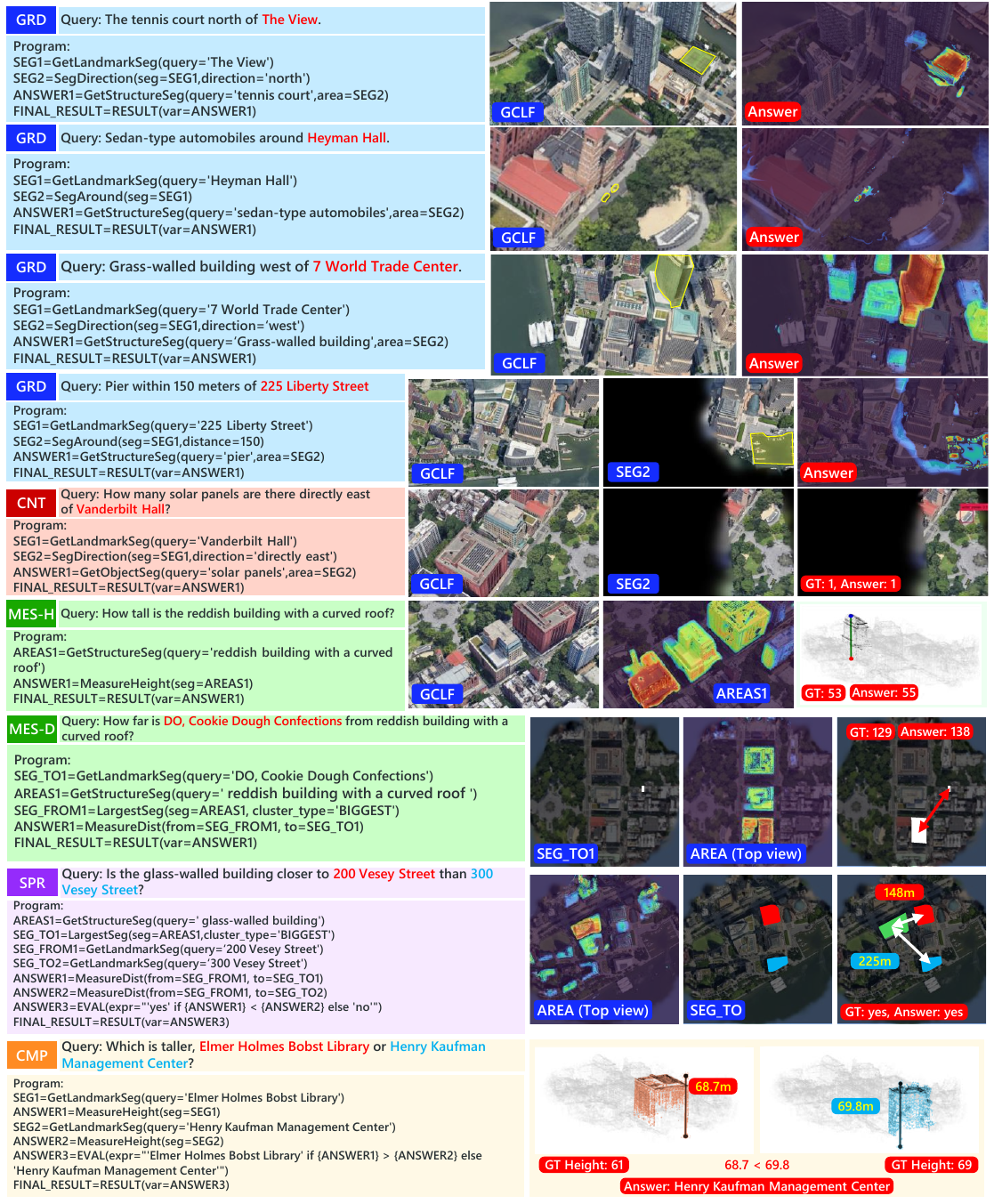}
\figvspacetop
\caption{Other qualitative results.}
\label{fig:other_qualitative_results}
\figvspace
\vspace{4pt}
\end{figure*}

\begin{figure*}
\centering
\includegraphics[scale=1.1]{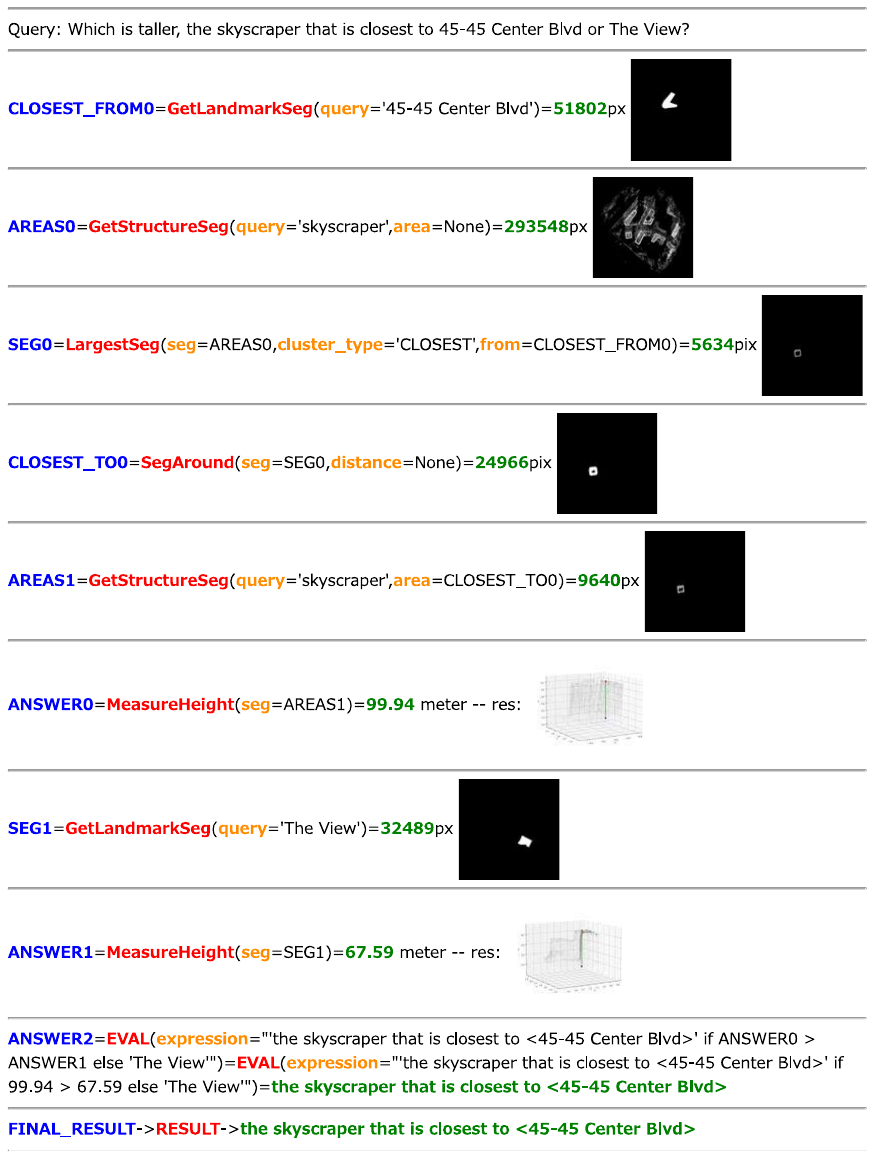}
\figvspacetop
\caption{Visual rationales generated by GeoProg3D.}
\label{fig:visprog_output}
\figvspace
\vspace{4pt}
\end{figure*}

\begin{figure*}
\centering
\includegraphics[width=\linewidth]{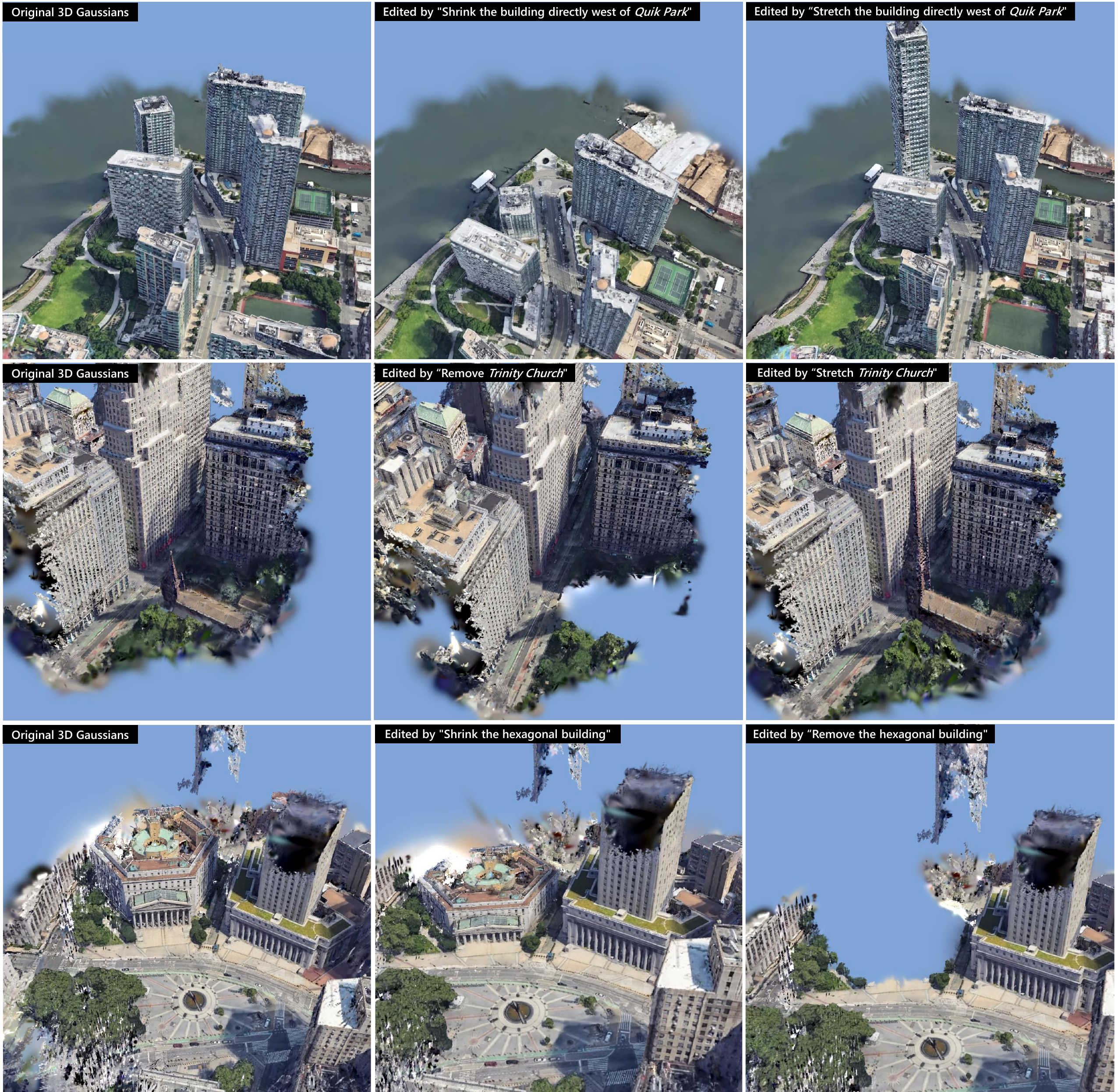}
\figvspacetop
\caption{Examples of language guided 3D Gaussian editing.}
\label{fig:visprog_editing}
\figvspace
\vspace{4pt}
\end{figure*}

\noindent \textbf{Qualitative examples.}
Figure~\ref{fig:other_qualitative_results} shows additional qualitative examples, demonstrating the capability of GeoProg3D across various tasks and environments. 
Figure~\ref{fig:visprog_output} illustrates the output obtained by executing a notebook included with our code.
Figure~\ref{fig:visprog_editing} shows examples of language-guided 3D Gaussian editing as an additional task.
This editing task requires models to localize the object and modify it based on a given query $q_k$.

\section{Dataset details}
\label{sec:ap_dataset_details}
To further ensure the quality and reliability of the dataset, we evaluated the inter-annotator agreement. For the SPR task, annotations from two independent annotators showed substantial agreement with a Cohen's Kappa of 0.78. For the MES-H and MES-D tasks, we observed high agreement as well, with Pearson correlation coefficients of 0.81 and 0.95, respectively. These results confirm that our annotations are consistent and reliable. Furthermore, for the SPR task, we explicitly instructed annotators to maintain a balanced label distribution, resulting in approximately 54\% "yes" and 46\% "no" answers to avoid potential bias.

\end{document}